\documentclass{article}

\usepackage[dblblindworkshop, final]{neurips_2026}
\workshoptitle{2nd Workshop on Advances in Representation Learning for Earth Observation}

\usepackage{amsmath,amsfonts,bm}

\def\eqref#1{equation~\ref{#1}}
\def\1{\bm{1}}

\DeclareMathAlphabet{\mathsfit}{\encodingdefault}{\sfdefault}{m}{sl}
\SetMathAlphabet{\mathsfit}{bold}{\encodingdefault}{\sfdefault}{bx}{n}

\usepackage[T1]{fontenc}
\usepackage[utf8]{inputenc}
\usepackage{graphicx}
\usepackage{float}
\usepackage{subcaption}
\usepackage{url}
\usepackage{hyperref}
\usepackage{booktabs}
\usepackage{microtype}
\usepackage{xcolor}
\usepackage{fontawesome5}

\newcommand{\querytext}{\faIcon{comment}}
\newcommand{\queryimage}{\faIcon{image}}
\newcommand{\querygeo}{\faIcon{map-marker-alt}}

\title{EarthEmbeddingExplorer: A Web Application for Cross-Modal Retrieval of Global Satellite Images}

\author{
Yijie Zheng$^{1}$\thanks{\texttt{zhengyijie23@mails.ucas.ac.cn}. Part of the v3 update was done during the author's internship at ModelScope.} \hspace{0.9em}
Weijie Wu$^{1}$ \hspace{0.9em}
Zhuoyao He$^{1}$ \hspace{0.9em}
Bingyue Wu$^{1}$ \hspace{0.9em}
Long Zhao$^{2}$  \\
\bf Guoqing Li$^{2}$ \hspace{0.9em}
Mikolaj Czerkawski$^{3}$ \hspace{0.9em}
Konstantin Klemmer$^{4,5}$
\\[1.5em]
\small \textsuperscript{1}University of Chinese Academy of Sciences \hspace{0.9em}
\small \textsuperscript{2}AIRCAS \\
\small \textsuperscript{3}Asterisk Labs \hspace{0.9em}
\small \textsuperscript{4}LGND AI, Inc. \hspace{0.9em}
\small \textsuperscript{5}University College London
}

\begin{document}

\maketitle

\begin{abstract}
Earth observation foundation models and global embedding datasets are increasingly available, yet comparing heterogeneous representations and using them without substantial local infrastructure remain difficult. We present EarthEmbeddingExplorer, an open web application that unifies text-, image-, location-, and mixed-query retrieval across eight EO embedding models and supports qualitative cross-model analysis through global similarity maps and top-ranked examples. We provide eight model-specific embedding datasets for 248,719 globally distributed Sentinel-2 patches, each linked to its source observation for on-demand imagery access. EarthEmbeddingExplorer is deployed on ModelScope with cloud GPU runtime, enabling end-to-end exploration without local setup or GPU hardware. The released Earth embeddings and web applications are available at \url{https://modelscope.cn/collections/Major-TOM/Core-S2L2A-249k}. 
\end{abstract}

\section{Introduction}
Earth observation (EO) foundation models increasingly represent satellite imagery as reusable embeddings for retrieval, clustering, and downstream analysis. Large embedding collections now provide an efficient interface to global EO archives~\citep{czerkawski2024global,klemmer2025earth,fang2026earth}. Yet these representations differ in supervision, RGB or multispectral inputs, and use of text, location, or acquisition-time metadata, complicating comparison under a common setting.

% Earth observation (EO) foundation models increasingly represent satellite imagery as reusable embeddings for retrieval, clustering, and downstream analysis. Large embedding collections now provide an efficient interface to global EO archives~\citep{czerkawski2024global,klemmer2025earth,fang2026earth}. Yet these representations use different supervision signals, operate on RGB or multispectral inputs, and may incorporate text, location, or acquisition-time metadata, making comparison under a common setting difficult.

Each model typically requires its own checkpoints, preprocessing, metadata conventions, and embedding index. Users must also connect retrieved vectors back to source imagery and implement geographic visualization. Consequently, differences in the semantic, visual, spatial, and temporal preferences of published embeddings remain difficult to examine.

We present \textbf{EarthEmbeddingExplorer}, an open interactive system for cross-modal retrieval and qualitative diagnosis of global Earth embeddings. It provides a common interface to eight models spanning vision--language, self-supervised, location-aware, multispectral, and spatiotemporal representations. The system supports text, image, location, and mixed queries, side-by-side model comparison, global similarity maps, and result export. GeoParquet metadata links each embedding to its source Sentinel-2 observation, enabling RGB or multispectral imagery to be fetched on demand without downloading the full archive.

Our contributions are threefold: (i) an open web application that unifies text-, image-, location-, and mixed-query retrieval across eight heterogeneous EO embedding models and supports qualitative cross-model analysis through geographic distributions and top-ranked examples; (ii) eight model-specific embedding datasets for 248,719 globally distributed Sentinel-2 patches, each linked to its source observation; and (iii) a cloud-hosted deployment on ModelScope with GPU runtime, enabling users to run the application without local setup or GPU hardware.

\section{EarthEmbeddingExplorer}
\subsection{Embedding models}
EarthEmbeddingExplorer currently includes eight complementary embedding models (Table~\ref{tab:models}) to support different query modes and comparison needs. FarSLIP~\citep{li2025farslip}, SigLIP~\citep{zhai2023sigmoid}, and TIPSv2~\citep{cao2026tipsv2} enable \textit{text-to-image} retrieval; DINOv2~\citep{oquab2024dinov}, Clay v1.5~\citep{clay2024foundation}, and OlmoEarth v1.2~\citep{tseng2026olmoearth} provide complementary visual, multispectral, and spatiotemporal representations; SatCLIP~\citep{klemmer2025satclip} enables \textit{location-to-image} retrieval; and Qwen3-VL-Embedding~\citep{li2026qwen3vlembedding} additionally supports native mixed text--image queries. All models support image queries, enabling users to contrast semantic alignment against visual and spatiotemporal similarity in a unified interface.

\subsection{Embedding datasets}
We utilize MajorTOM-Core-S2L2A~\citep{francis2024major} as the imagery source. The dataset is indexed via a systematic grid of approximately $10 \times 10$~km cells, ensuring comprehensive spatial coverage. We uniformly select one out of every nine grid cells and extract one central $384 \times 384$ pixel patch from each selected source tile. Thus, the $1/9$ ratio describes grid-cell sampling before cropping, rather than the fraction of source pixels retained. This process yields 248,719 unique patches, representing approximately 1.4\% of Earth's land surface (Figure~\ref{fig:samples}). Following the Major~TOM Embedding Expansions standard~\citep{czerkawski2024global}, we release these as precomputed embeddings stored in GeoParquet shards. This cloud-native format enables high-speed lookups and efficient partial downloads, which are essential for real-time interactive visualization in our web application.

\begin{table}[t]
\centering
\small
\setlength{\tabcolsep}{3.5pt}
\renewcommand{\arraystretch}{1.08}
\resizebox{\linewidth}{!}{%
\begin{tabular}{lllll}
\hline
\textbf{Model / variant}
& \textbf{Training data}
& \textbf{Encoder input}
& \textbf{Query}
& \textbf{Dim.} \\
\hline

FarSLIP (ViT-B/16)
& RS5M, MGRS-200K
& 224$\times$224 RGB
& \querytext\enspace\queryimage
& 512 \\

Qwen3-VL-Embedding-2B
& Synthetic + curated multimodal
& Dynamic RGB
& \querytext\enspace\queryimage
& 2,048 \\

SigLIP (ViT-SO400M/14)
& WebLI
& 384$\times$384 RGB
& \querytext\enspace\queryimage
& 1,152 \\

TIPSv2 (ViT-B/14)
& WebLI-116M + synthetic
& 448$\times$448 RGB
& \querytext\enspace\queryimage
& 768 \\

SatCLIP (ViT16-L40)
& S2-100K
& 224$\times$224, 12-band S2
& \queryimage\enspace\querygeo
& 256 \\

DINOv2-large (ViT-L/14)
& LVD-142M
& 224$\times$224 RGB
& \queryimage
& 1,024 \\

Clay v1.5 (ViT-L/8 MAE)
& 70M multi-sensor EO
& 384$\times$384, 10-band S2
& \queryimage
& 1,024 \\

OlmoEarth v1.2 Base (FlexiViT-B/8)
& S1/S2/L8 + 6 maps
& 128$\times$128, 12-band S2
& \queryimage
& 768 \\

\hline
\end{tabular}%
}
\caption{Embedding models used in EarthEmbeddingExplorer. Model input denotes
the resolution and spectral channels supplied to the encoder. Clay and
OlmoEarth additionally use acquisition metadata, while Qwen3-VL-Embedding
supports dynamic-resolution RGB inputs and native mixed text--image queries.
Query modalities are denoted by \querytext~text, \queryimage~image, and
\querygeo~geolocation.}
\label{tab:models}
\end{table}

\begin{figure}[H]
    \centering
    \includegraphics[width=0.99\linewidth]{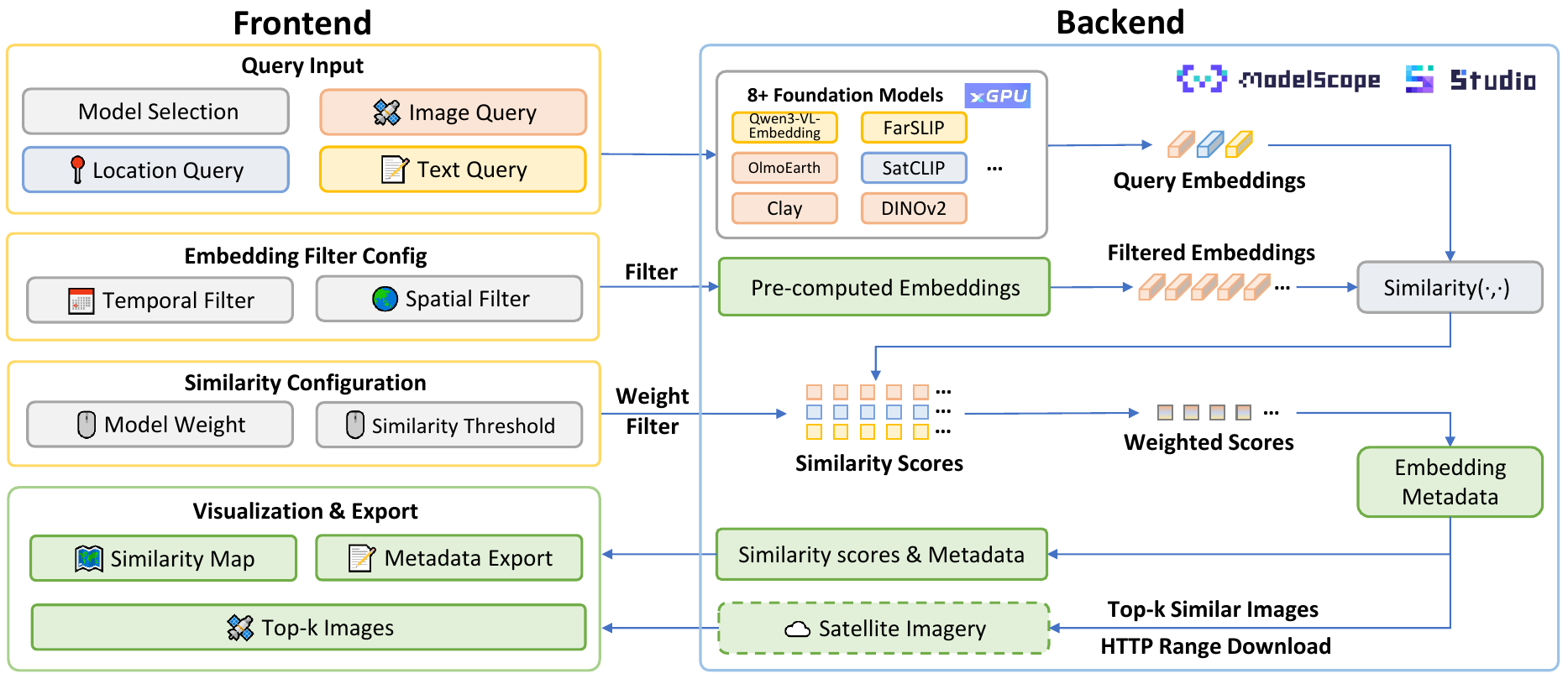}
    \caption{A cloud-native retrieval pipeline based on ModelScope Studio.}
    \label{fig:architecture}
\end{figure}
\subsection{System architecture}
\begin{figure}[H]
    \centering
    \includegraphics[width=0.98\linewidth]{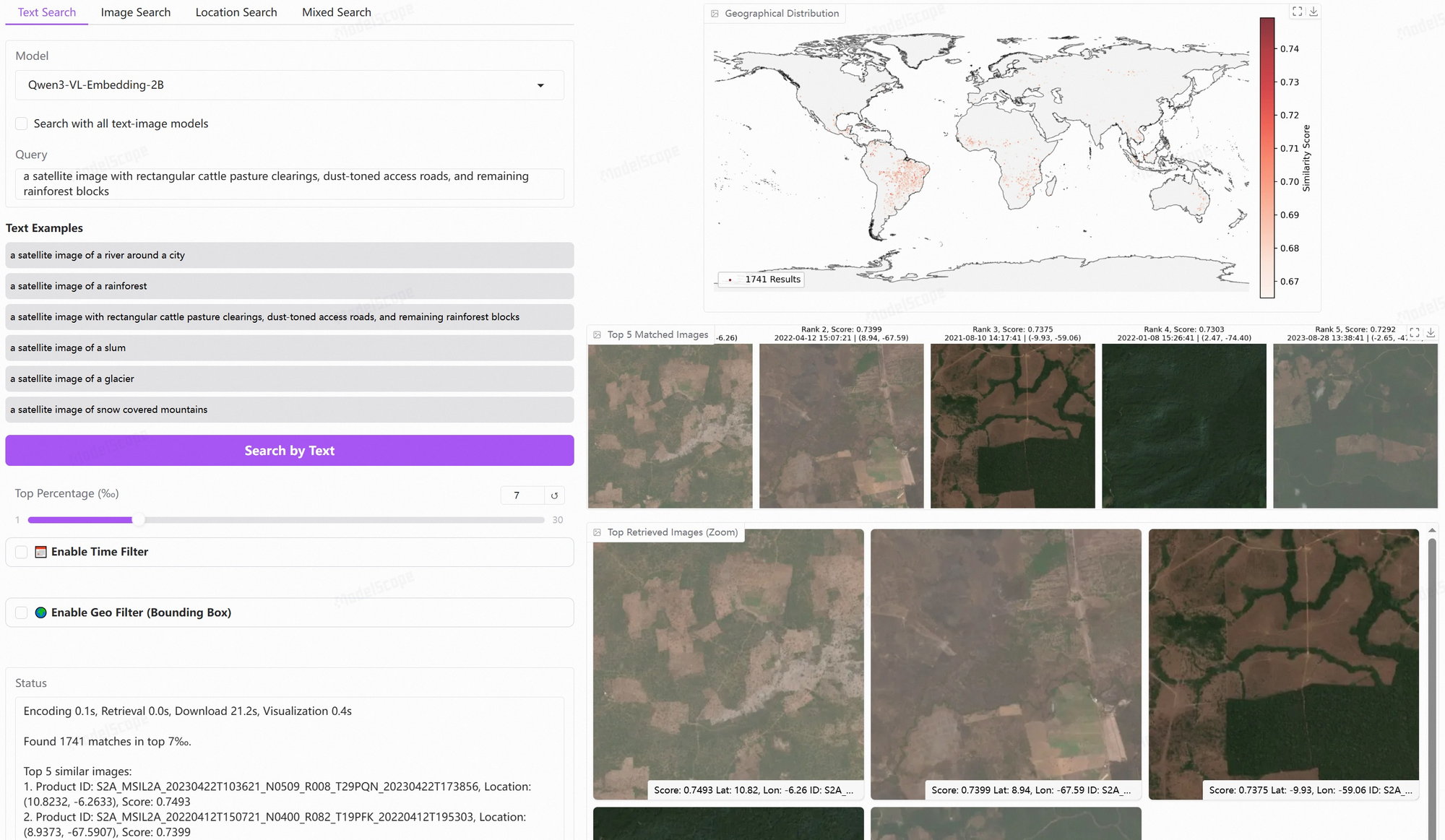}
    \caption{EarthEmbeddingExplorer user interface.}
    \label{fig:ui}
\end{figure}

Figure~\ref{fig:architecture} summarizes the cloud-native design. Queries are embedded with selected models, matched via vector similarity search over precomputed embeddings, and visualized as a similarity map and top-$k$ retrieved images. We offer ModelScope deployments with free GPU runtime, allowing users to access the application without local setup. The WebAPP is built with Gradio~\citep{abid2019gradio}. As shown in Figure~\ref{fig:ui}, the left panel configures inputs, while the right panel visualizes similarity maps and retrieved examples.

% Choose representative models with different training data. SigLIP is a well-known language-image pretraining model trainied on natural image, while FarSLIP is the first Finegrained Aligned CLIP style model in remote sensing. DINOv2 is , SatCLIP... The four models are representatives of different input modality, pretraining method, and training data domain.
% supporting text-to-image (SigLIP, FarSLIP), location-to-image (SatCLIP), and image-to-image (all) retrieval
% % Model	Encoder type	Training data Support mode
% % SigLIP	image encoder + text encoder	natural image–text pairs from the web
% % DINOv2	image encoder only	web-scale natural images (self-supervised)
% % FarSLIP	image encoder + text encoder	satellite image–text pairs
% % SatCLIP	image encoder + location encoder	satellite image–location pairs

% Standard follow Major TOM Embedding Datasets released under Major TOM embedding stand, Geoparquet file, Metadata including
% % | Column | unique_id | embedding | timestamp | product_id | grid_cell | grid_row_u | grid_col_r | geometry | centre_lat | centre_lon | utm_footprint | utm_crs | pixel_bbox | fields

\begin{figure}[h]
    \centering
    \begin{subfigure}{0.48\textwidth}
        \centering
        \includegraphics[width=\textwidth]{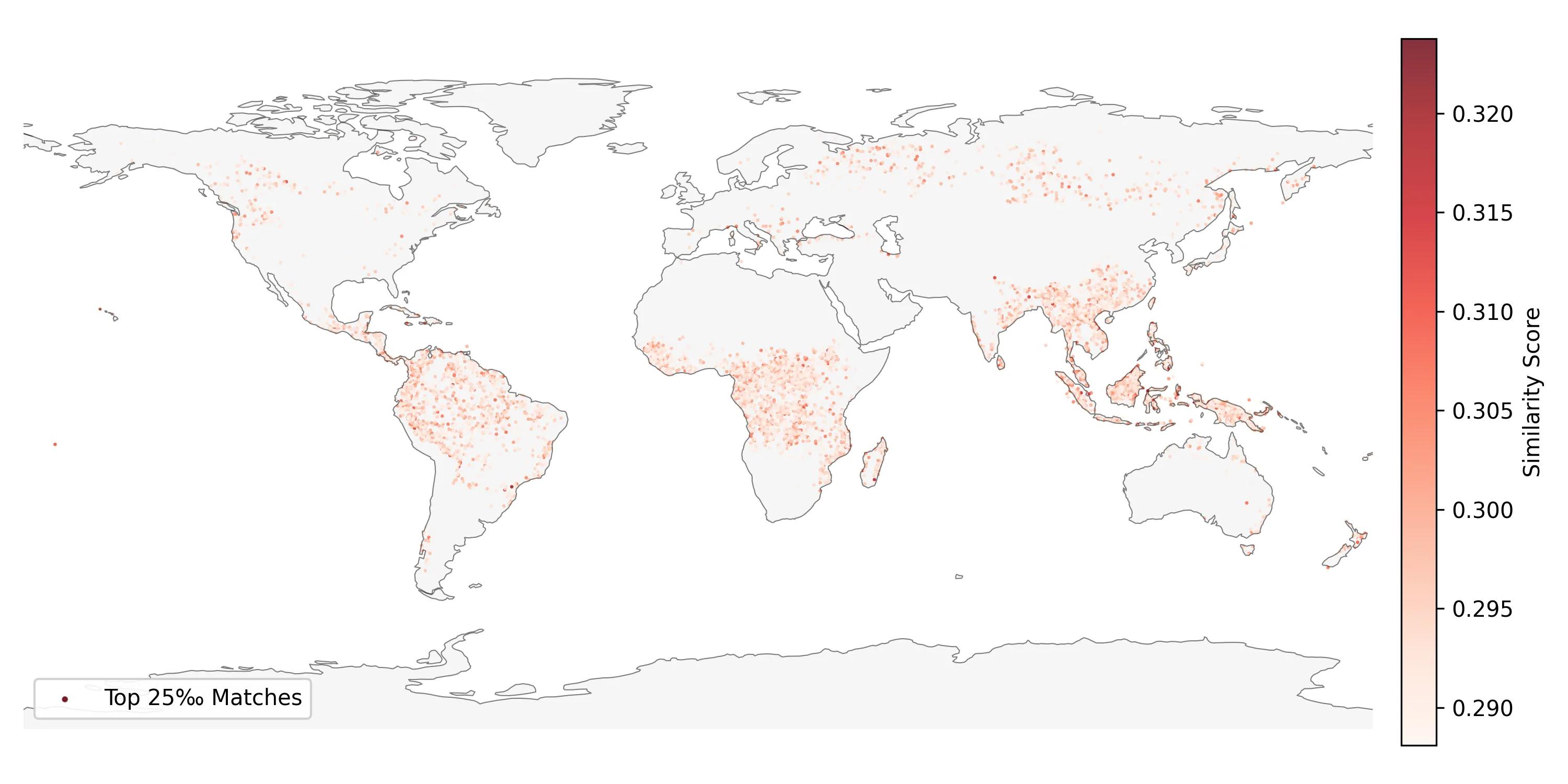}
        \caption{FarSLIP, text query}
    \end{subfigure}
    \hfill
    \begin{subfigure}{0.48\textwidth}
        \centering
        \includegraphics[width=\textwidth]{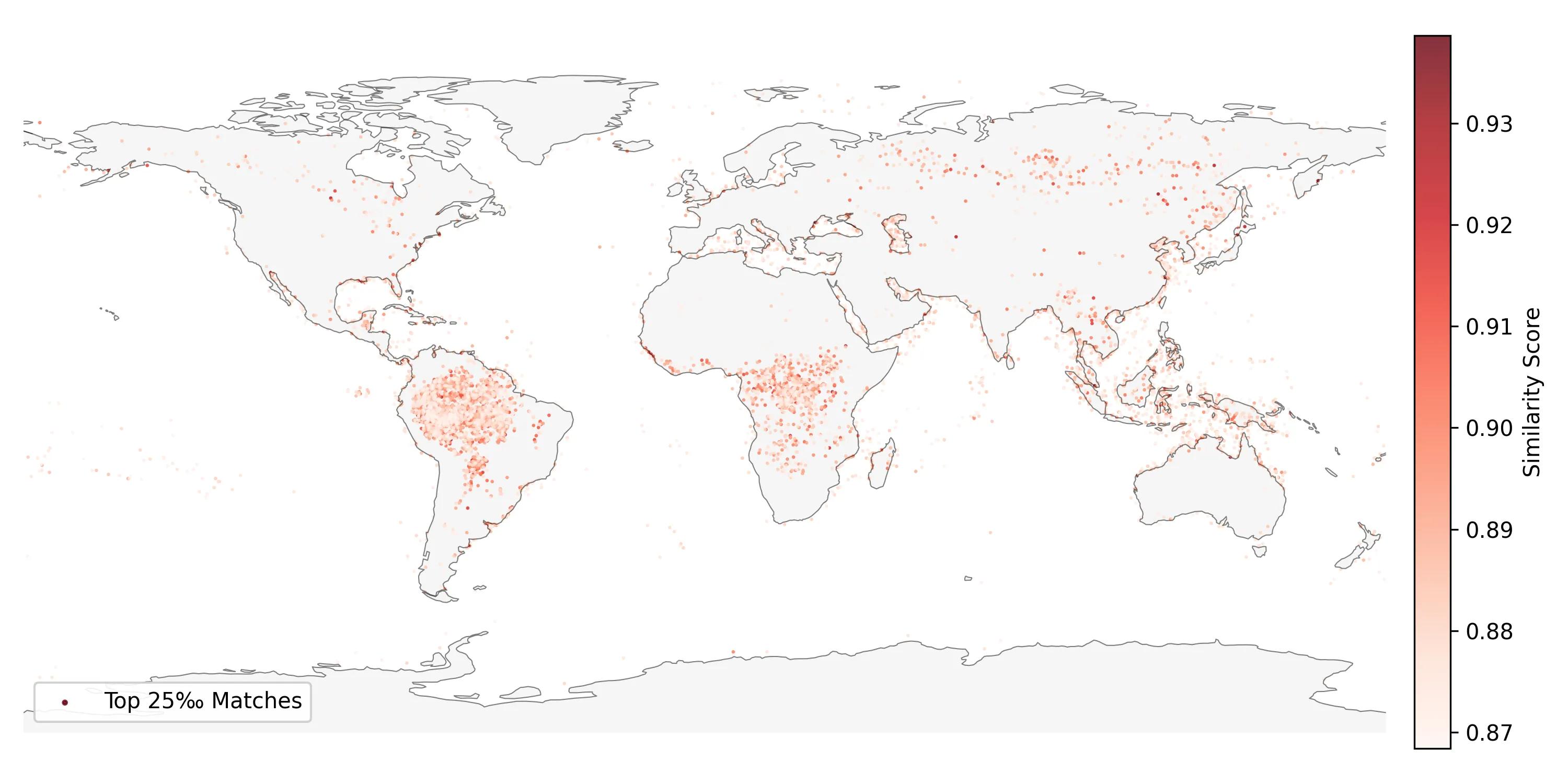}
        \caption{FarSLIP, image query}
    \end{subfigure}
    \\
    \begin{subfigure}{0.48\textwidth}
        \centering
        \includegraphics[width=\textwidth]{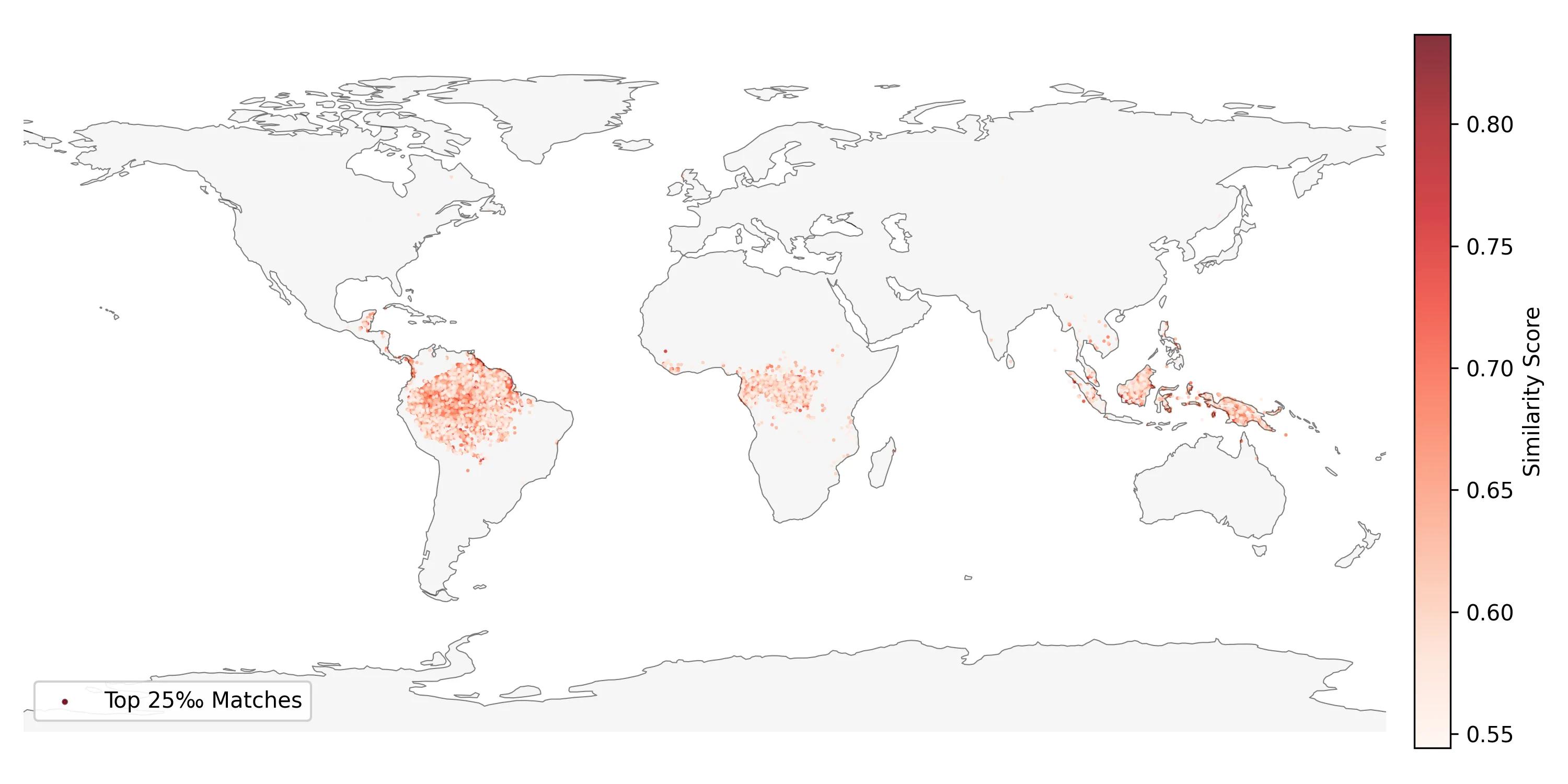}
        \caption{SatCLIP, location query}
    \end{subfigure}
    \hfill
    \begin{subfigure}{0.48\textwidth}
        \centering
        \includegraphics[width=\textwidth]{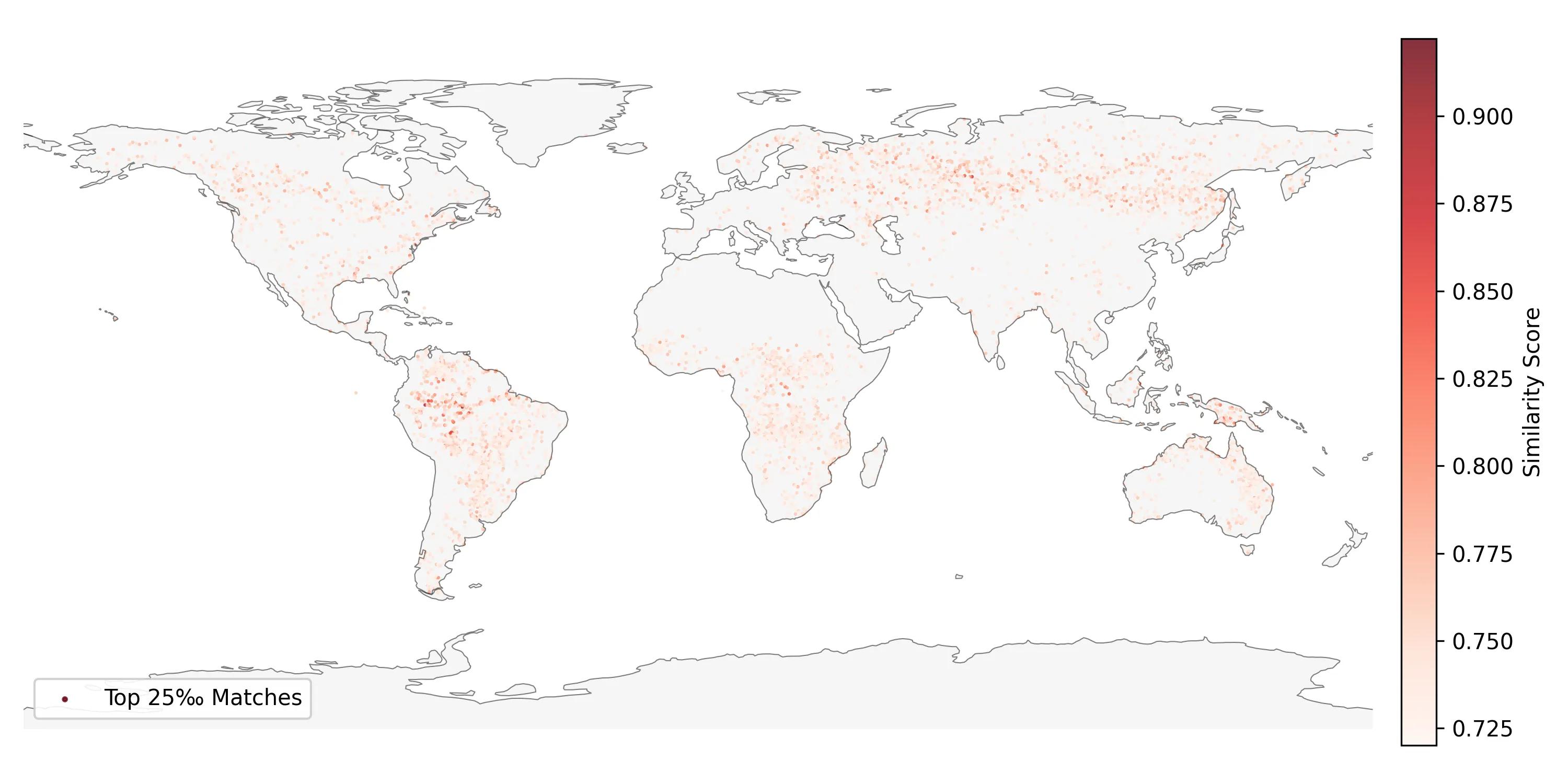}
        \caption{DINOv2, image query}
    \end{subfigure}
    \caption{Geographic distribution of retrieved matches under a top-2.5\% threshold for different models and query modalities.}
    \label{fig:comparison_similarity_distribution}
\end{figure}

\section{Application and Qualitative Analysis}

We qualitatively analyze a representative rainforest application across query modalities by comparing global similarity ``hotspots'' and top-ranked matches. Additional text-to-image and image-to-image comparisons are provided in the Appendix.

\begin{figure}[h]
    \centering
    \begin{subfigure}{0.4\textwidth}
        \centering
        \includegraphics[width=\textwidth]{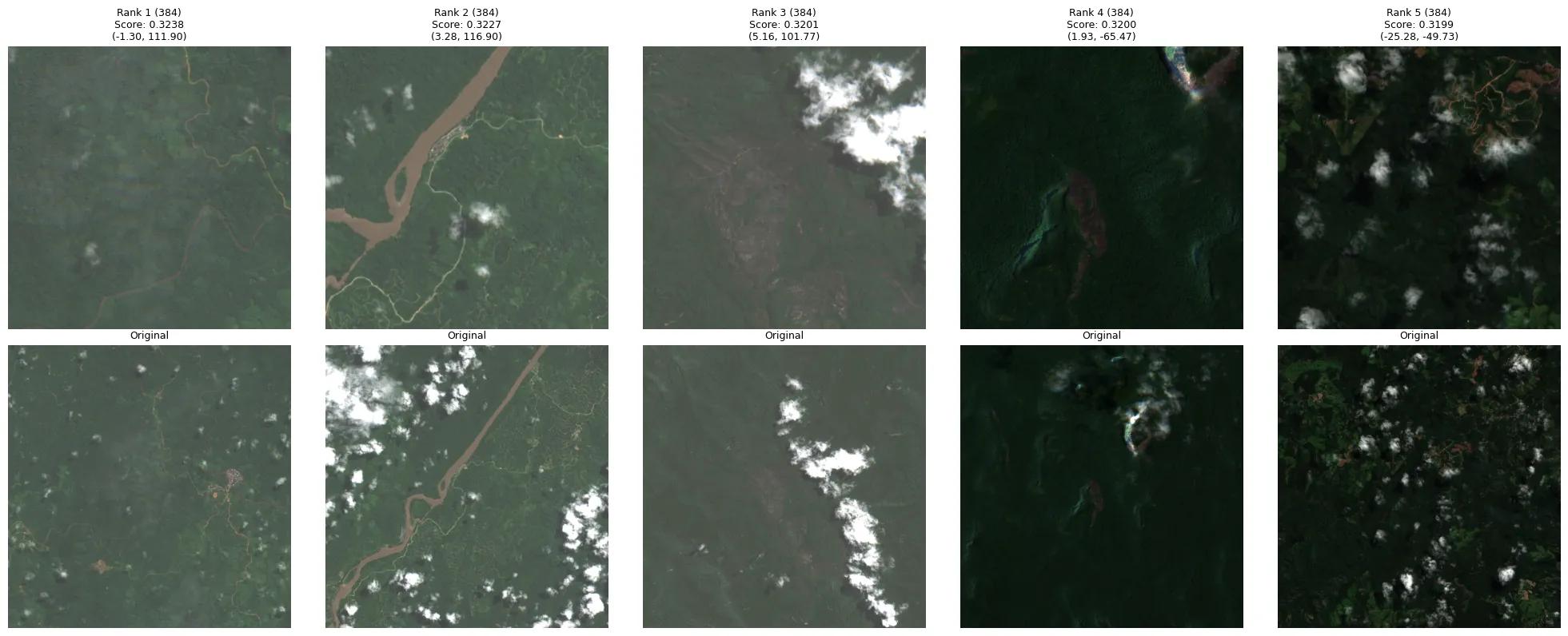}
        \caption{FarSLIP, text query}
    \end{subfigure}
    \hfill
    \begin{subfigure}{0.48\textwidth}
        \centering
        \includegraphics[width=\textwidth]{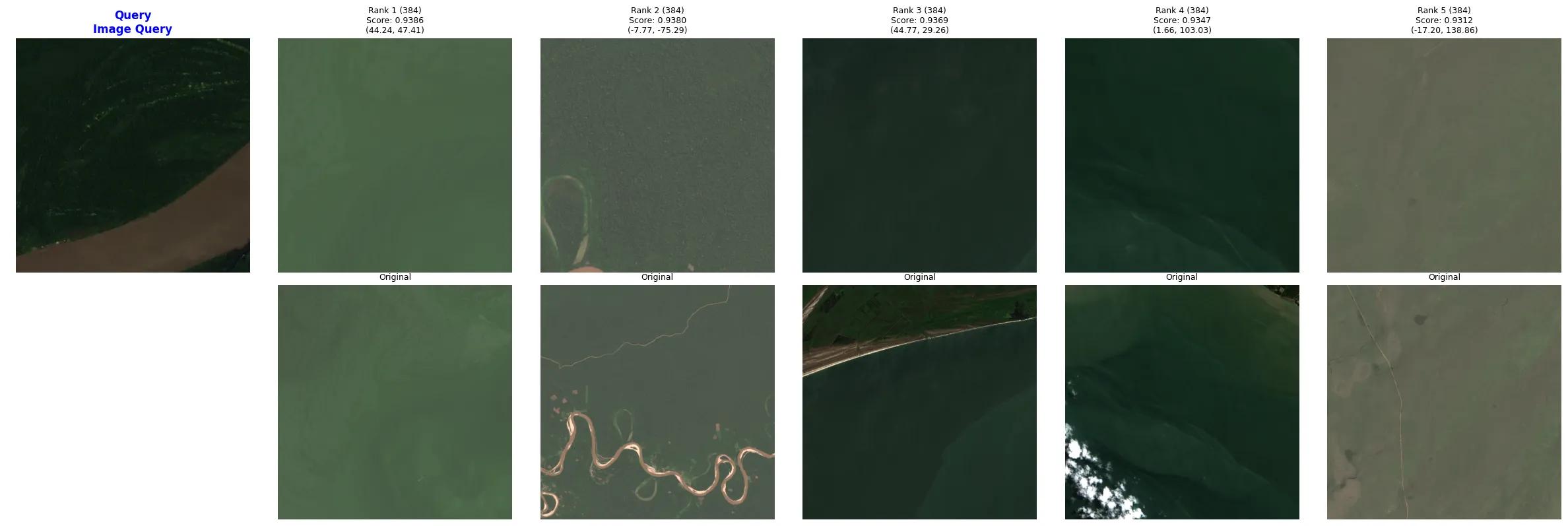}
        \caption{FarSLIP, image query}
    \end{subfigure}
    \\
    \begin{subfigure}{0.48\textwidth}
        \centering
        \includegraphics[width=\textwidth]{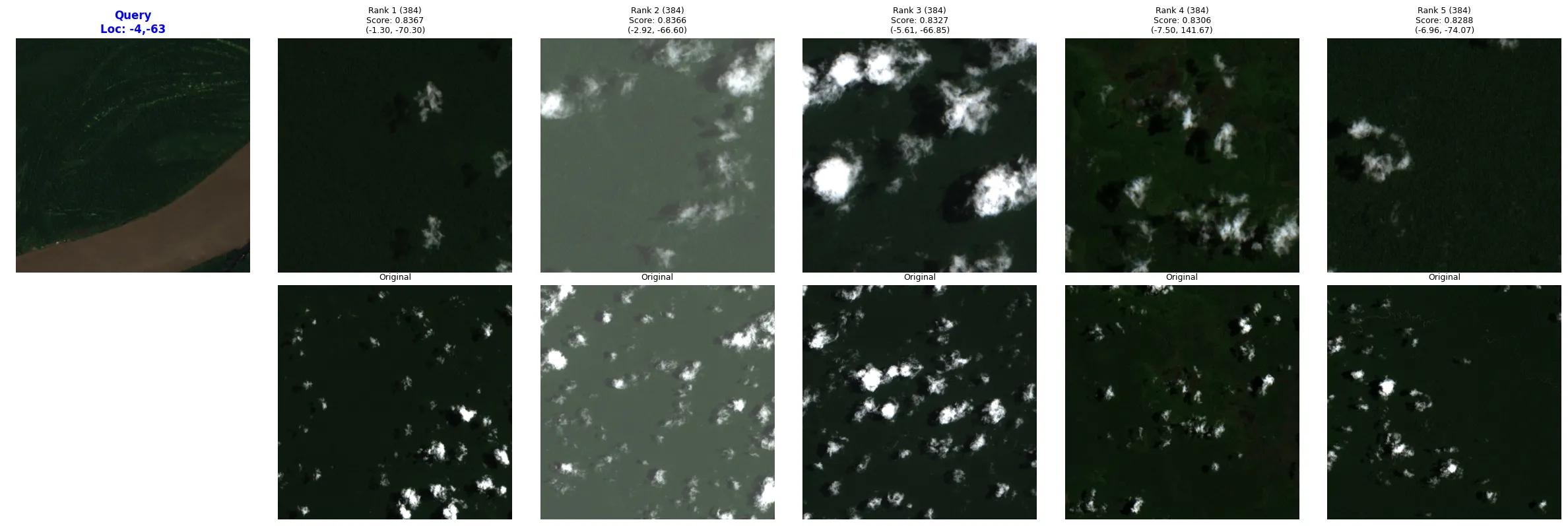}
        \caption{SatCLIP, location query}
    \end{subfigure}
    \hfill
    \begin{subfigure}{0.48\textwidth}
        \centering
        \includegraphics[width=\textwidth]{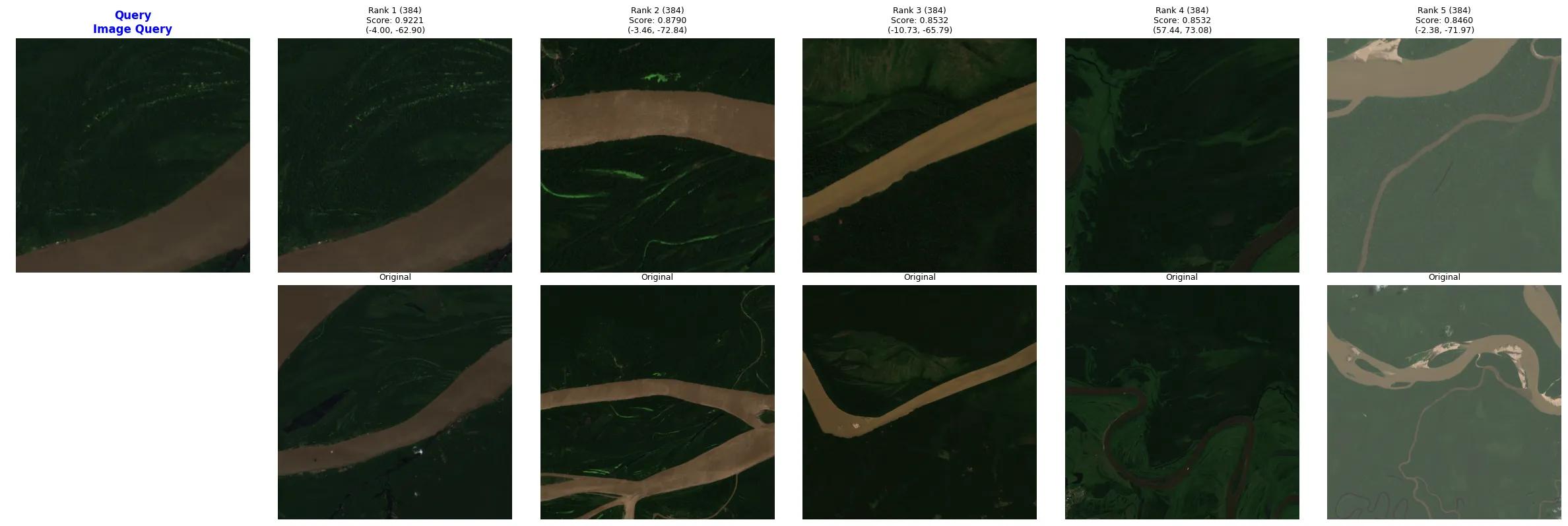}
        \caption{DINOv2, image query}
    \end{subfigure}
    \caption{Top-5 retrieved tiles for the same case study in Figure~\ref{fig:comparison_similarity_distribution}.}
    \label{fig:comparison_top5}
\end{figure}

We demonstrate the workflow with a rainforest retrieval case study.
For \textit{text-to-image} search, we use the prompt \texttt{a satellite image of a tropical rainforest}.
For \textit{image-to-image} search, the query is an image patch centered at (4$^\circ$S, 63$^\circ$W) near Rio Purus (an upstream tributary of the Amazon).
For \textit{location-to-image} search, we use the same coordinates as the location query.

Figure~\ref{fig:comparison_similarity_distribution} compares the geographic distribution of high-scoring matches. For the text query, FarSLIP concentrates matches in humid tropical regions, consistent with semantic alignment to \emph{rainforest}. SatCLIP scores are largely concentrated in the tropical belt, including the Amazon Basin, Congo Basin, and Southeast Asia. Image-query matches are more dispersed, possibly because visual patterns such as rivers and dark vegetation recur across climates and continents.

Figure~\ref{fig:comparison_top5} shows the top-5 patches for each setting. Text-based results are generally consistent with rainforest scenes and often include cloud cover. Four of DINOv2's top-5 results contain wide rivers, suggesting a strong contribution from river morphology. FarSLIP image retrieval instead returns rainforest-like patches less centered on the river pattern, illustrating a qualitative difference between semantic and visual similarity.

The results also expose limitations: FarSLIP occasionally scores patches outside the expected climate zone for ``tropical rainforest,'' suggesting limited geographic or climatic priors, while image retrieval produces occasional implausible matches such as ocean tiles. Further analyses of socioeconomic and natural-scene concepts are reported in the Appendix.

\section{Conclusions and Roadmap}
EarthEmbeddingExplorer packages precomputed Earth embeddings into an interactive, reproducible workflow for cross-modal retrieval and rapid qualitative evaluation. It is intended both for model developers (to stress-test representations at global scale) and for geoscience users (to quickly find and export regions of interest from flexible text/image/location queries).

Next steps are to expand spatial and temporal coverage and build a global retrieval benchmark for quantitative comparison of model behavior and performance. We will also support community-driven additions of embeddings and models.

\begin{ack}
This work was partly supported by 4th IEEE GRSS Student Grand Challenge. We thank ModelScope for providing free access to the high-performance GPU resources used to deploy the web application.
\end{ack}

% The maximum length for submission is 4 pages excluding references. You may add additional information into the supplementary appendix section after the references.

% Please make sure that the submitted paper is anonymized!

\bibliography{references}
\bibliographystyle{plainnat}

\newpage
\appendix
\section{Appendix}
\begin{figure}[H]
    \centering
    \includegraphics[width=0.61\linewidth]{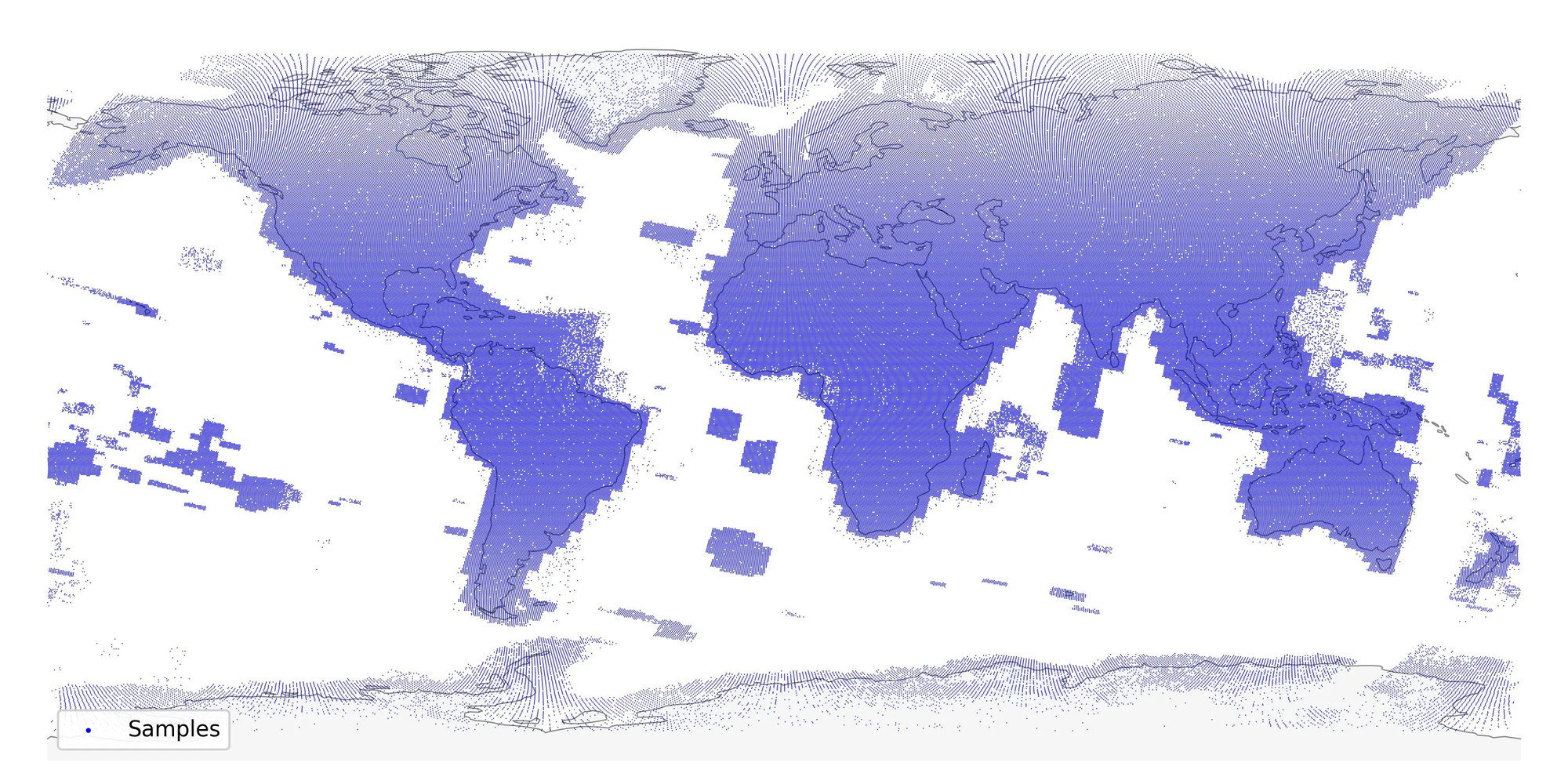}
    \caption{Geographical distribution of sampled grids}
    \label{fig:samples}
\end{figure}

This appendix provides supplementary details, including the geographical distribution of our sampled grids (Figure~\ref{fig:samples}) and additional qualitative cross-model analyses for text-to-image and image-to-image retrieval.

\subsection{Additional cross-model comparison for text-to-image retrieval}
We compare FarSLIP, SigLIP, TIPSv2, and Qwen3-VL-Embedding on text-to-image retrieval using prompts from representative Earth observation applications~\citep{lee2026generalizable,xie2022glaciernet2}. For each prompt, we jointly inspect the global distribution of high-scoring samples and the top-5 retrieved tiles. Similarity scores are model-specific and are therefore not compared across models.

\begin{figure}[H]
    \centering
    \begin{subfigure}{0.48\textwidth}
        \centering
        \includegraphics[width=\textwidth]{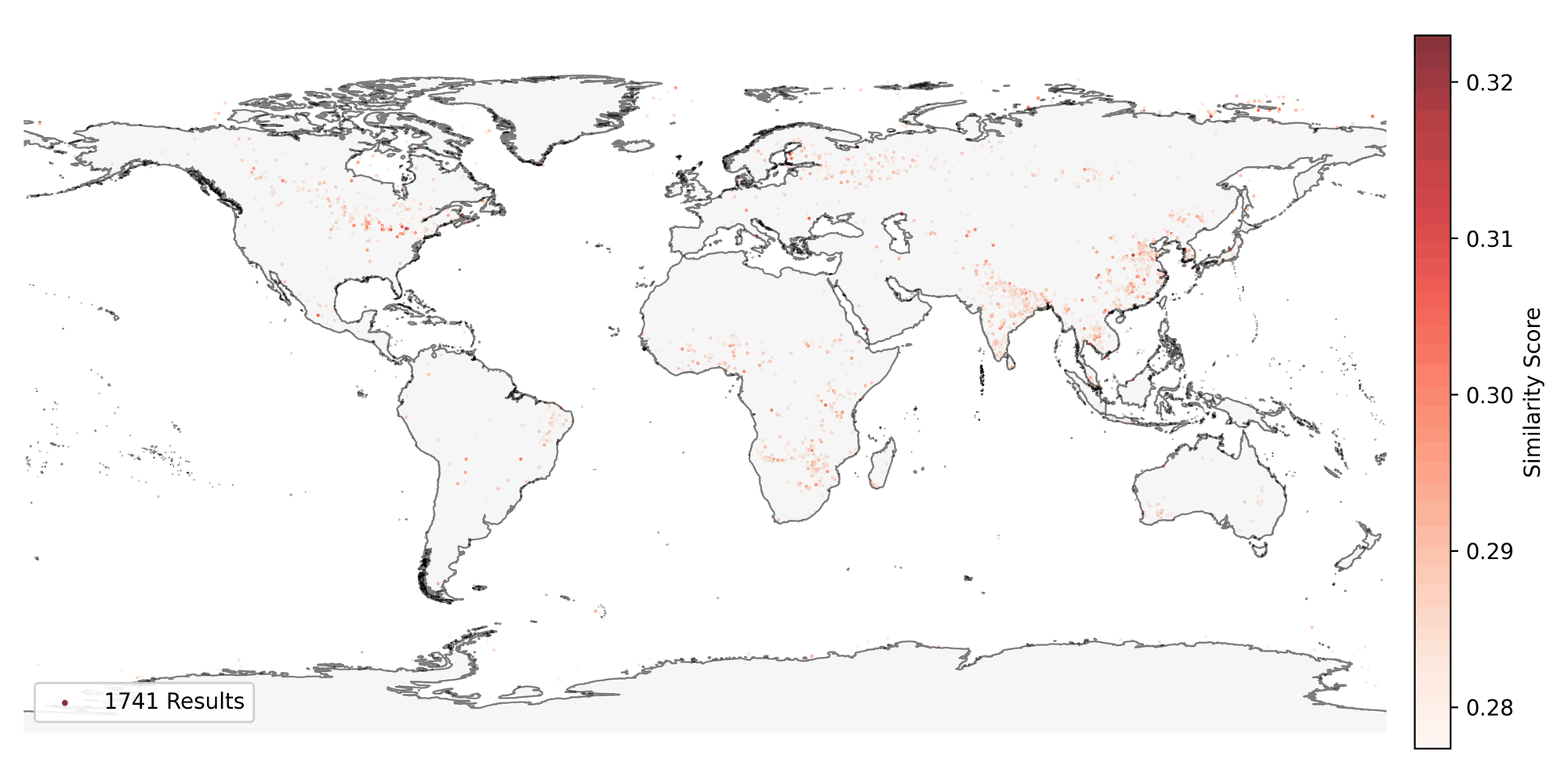}
        \caption{FarSLIP distribution}
    \end{subfigure}
    \hfill
    \begin{subfigure}{0.48\textwidth}
        \centering
        \includegraphics[width=\textwidth]{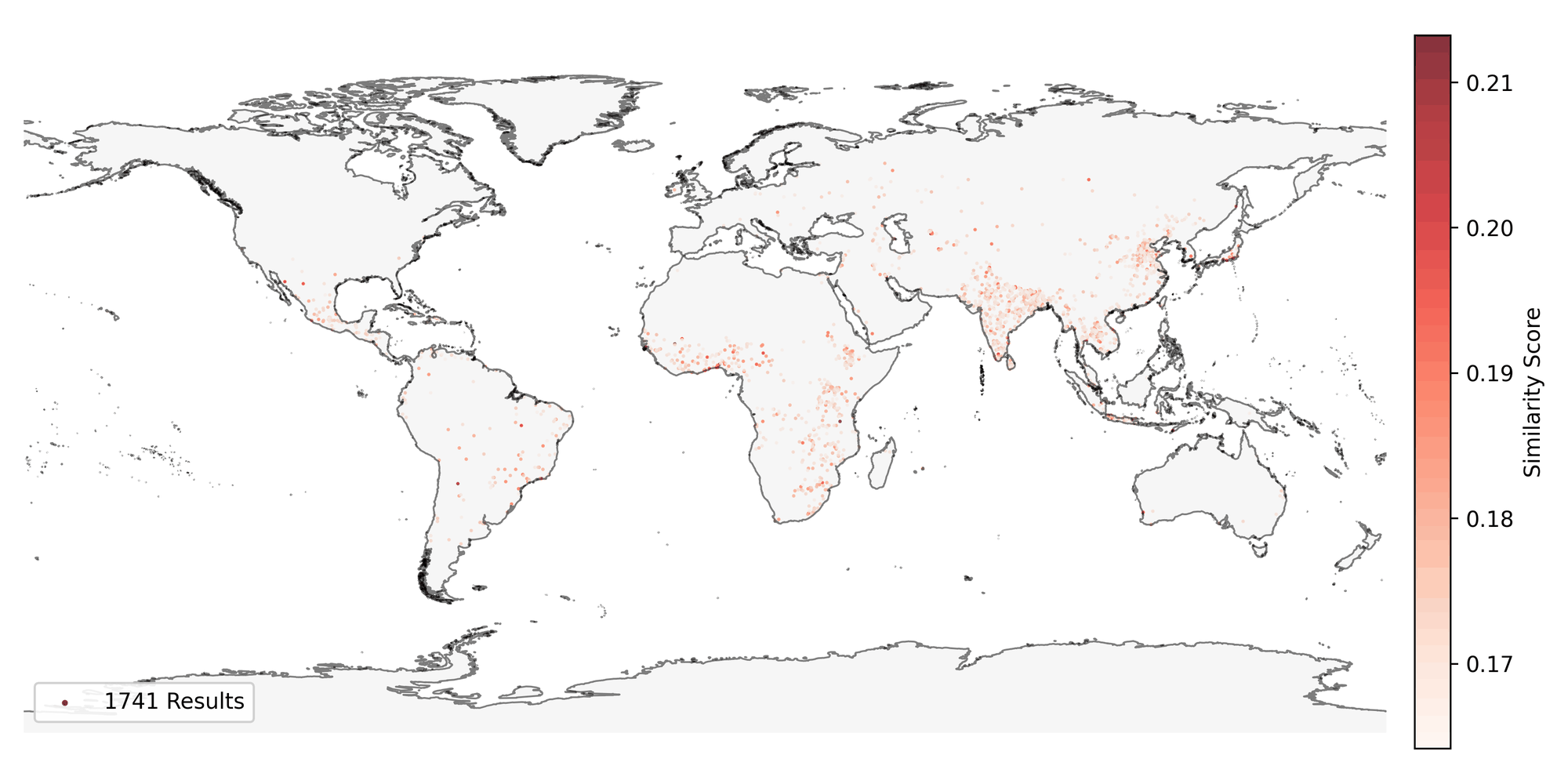}
        \caption{SigLIP distribution}
    \end{subfigure}
    \\
    \begin{subfigure}{0.48\textwidth}
        \centering
        \includegraphics[width=\textwidth]{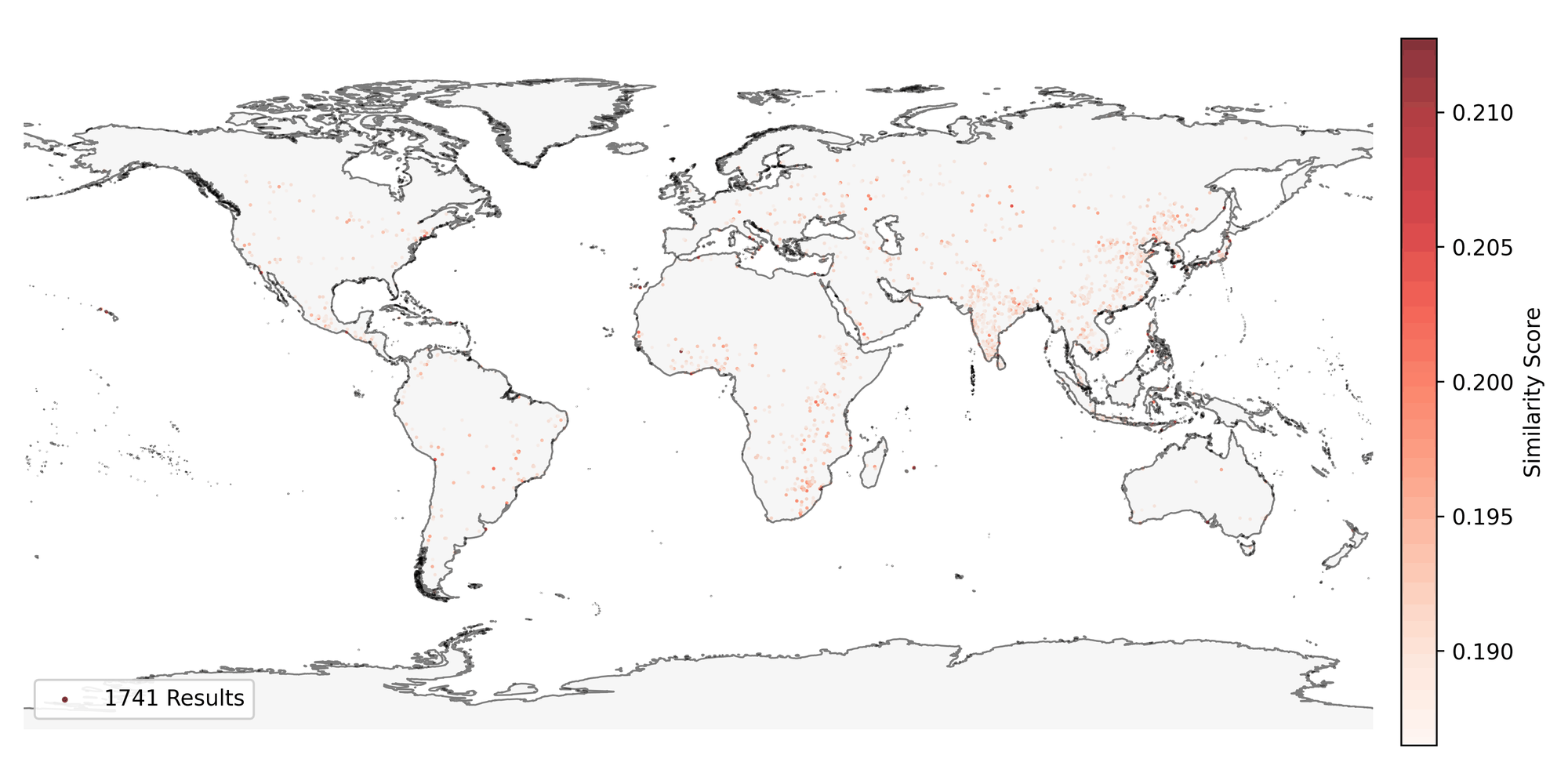}
        \caption{TIPSv2 distribution}
    \end{subfigure}
    \hfill
    \begin{subfigure}{0.48\textwidth}
        \centering
        \includegraphics[width=\textwidth]{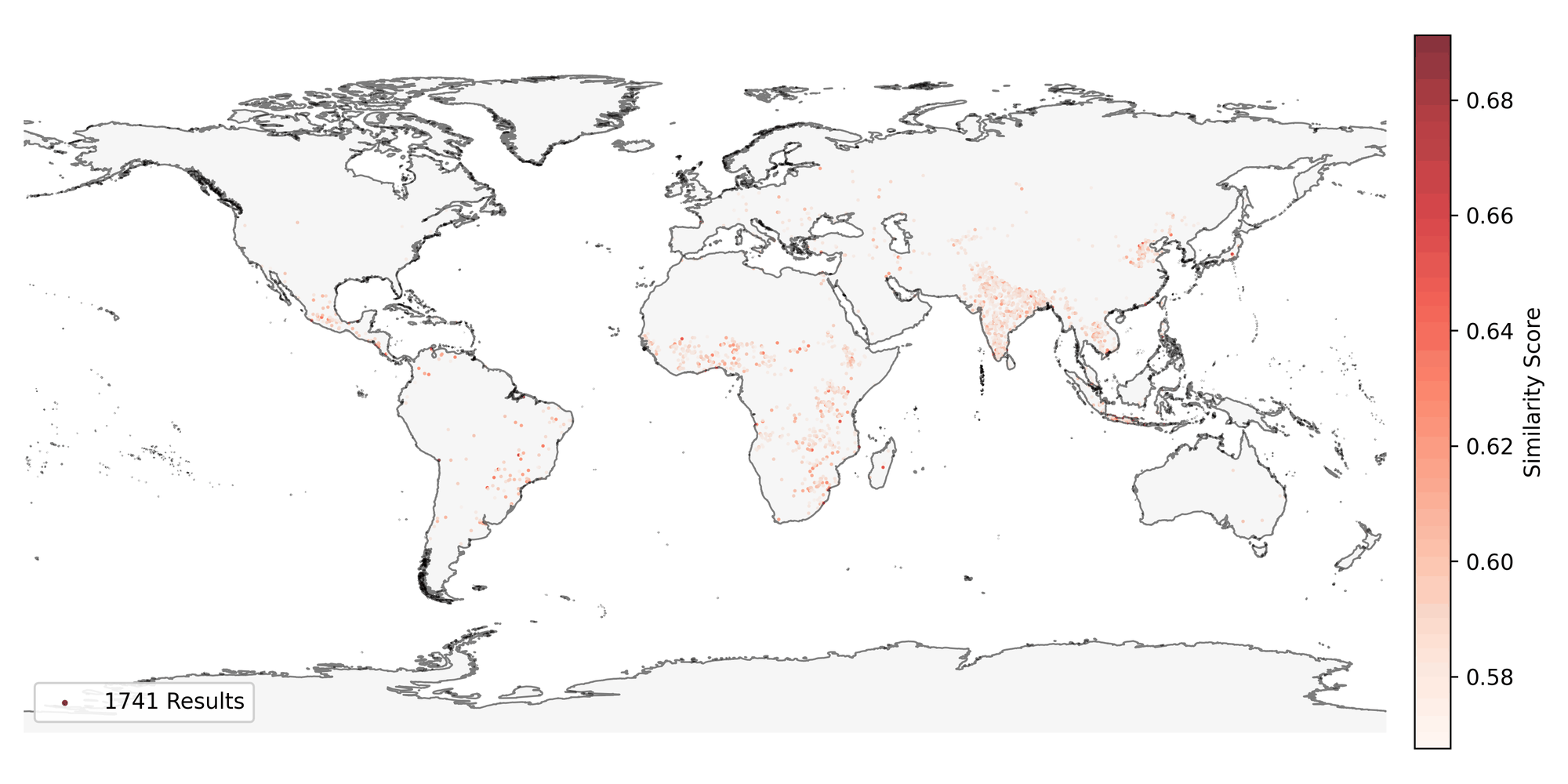}
        \caption{Qwen3-VL-Embedding distribution}
    \end{subfigure}
    \\
    \begin{subfigure}{0.48\textwidth}
        \centering
        \includegraphics[width=\textwidth]{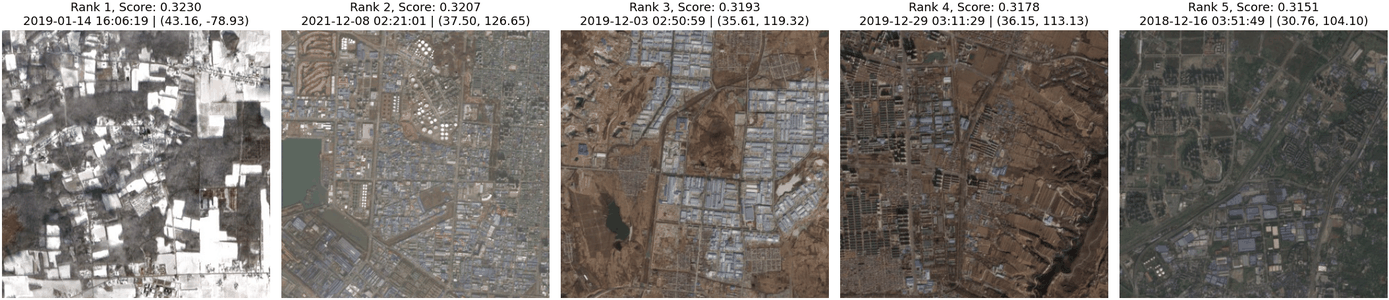}
        \caption{FarSLIP top-5}
    \end{subfigure}
    \hfill
    \begin{subfigure}{0.48\textwidth}
        \centering
        \includegraphics[width=\textwidth]{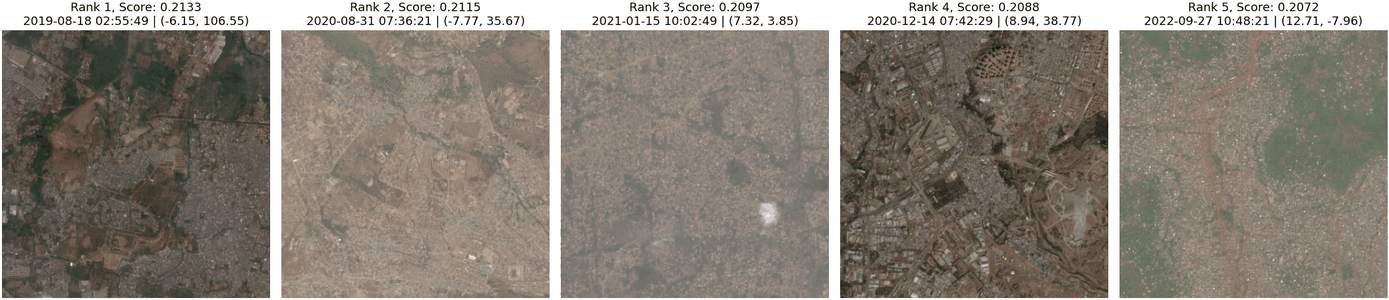}
        \caption{SigLIP top-5}
    \end{subfigure}
    \\
    \begin{subfigure}{0.48\textwidth}
        \centering
        \includegraphics[width=\textwidth]{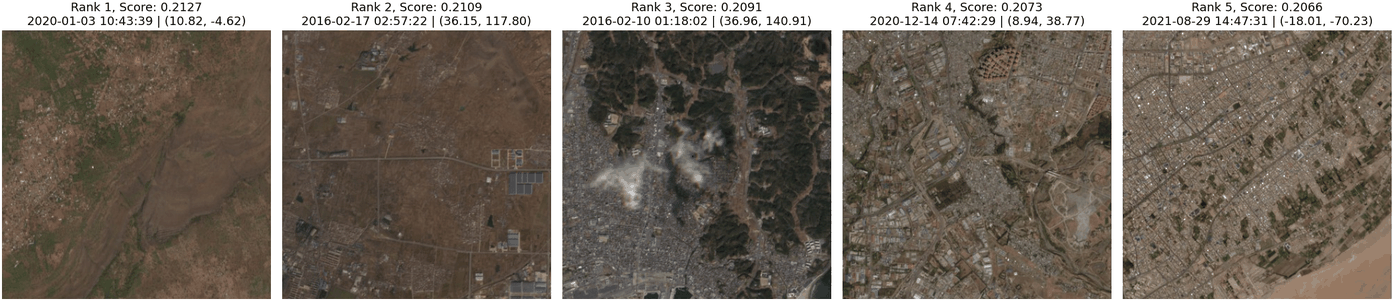}
        \caption{TIPSv2 top-5}
    \end{subfigure}
    \hfill
    \begin{subfigure}{0.48\textwidth}
        \centering
        \includegraphics[width=\textwidth]{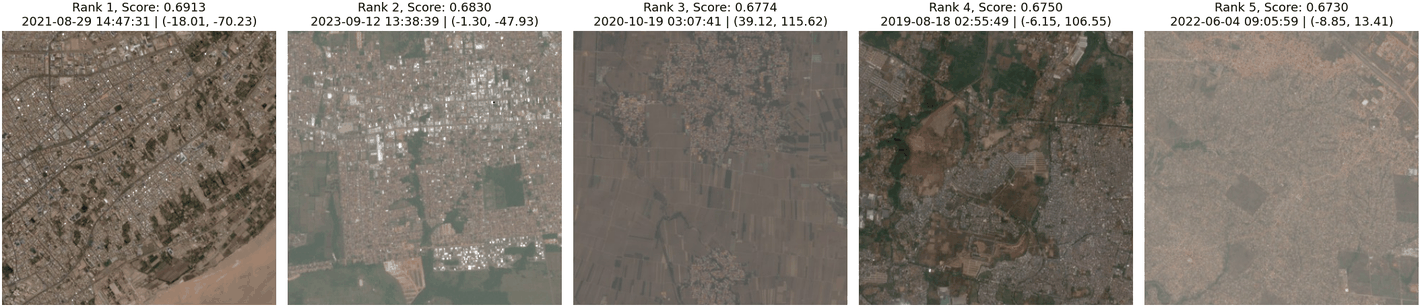}
        \caption{Qwen3-VL-Embedding top-5}
    \end{subfigure}
    \caption{Cross-model text-to-image retrieval for \texttt{a satellite image of a slum}. The first two rows show global distributions, and the last two rows show top-5 retrieved tiles.}
    \label{fig:slum}
\end{figure}

\clearpage
\paragraph{Socio-economic concept: slum}
Figure~\ref{fig:slum} compares results for \texttt{a satellite image of a slum}. SigLIP and Qwen3-VL-Embedding concentrate many high-scoring samples in South Asia, sub-Saharan Africa, and Latin America. Their top-ranked tiles predominantly show dense, fine-grained, low-rise built-up patterns, often along urban fringes. TIPSv2 exhibits a broader distribution and a more heterogeneous top-5, mixing dense settlements with sparsely developed, industrial, or cloud-affected scenes. FarSLIP is the most geographically diffuse, with responses extending across East Asia and North America; its top-5 includes dense urban or industrial textures as well as a snow-bright scene. These results suggest that the models rely on different visual proxies for this socio-economic concept. Accordingly, this case study compares learned retrieval behavior rather than slum-detection accuracy against ground-truth labels.

\begin{figure}[H]
    \centering
    \begin{subfigure}{0.48\textwidth}
        \centering
        \includegraphics[width=\textwidth]{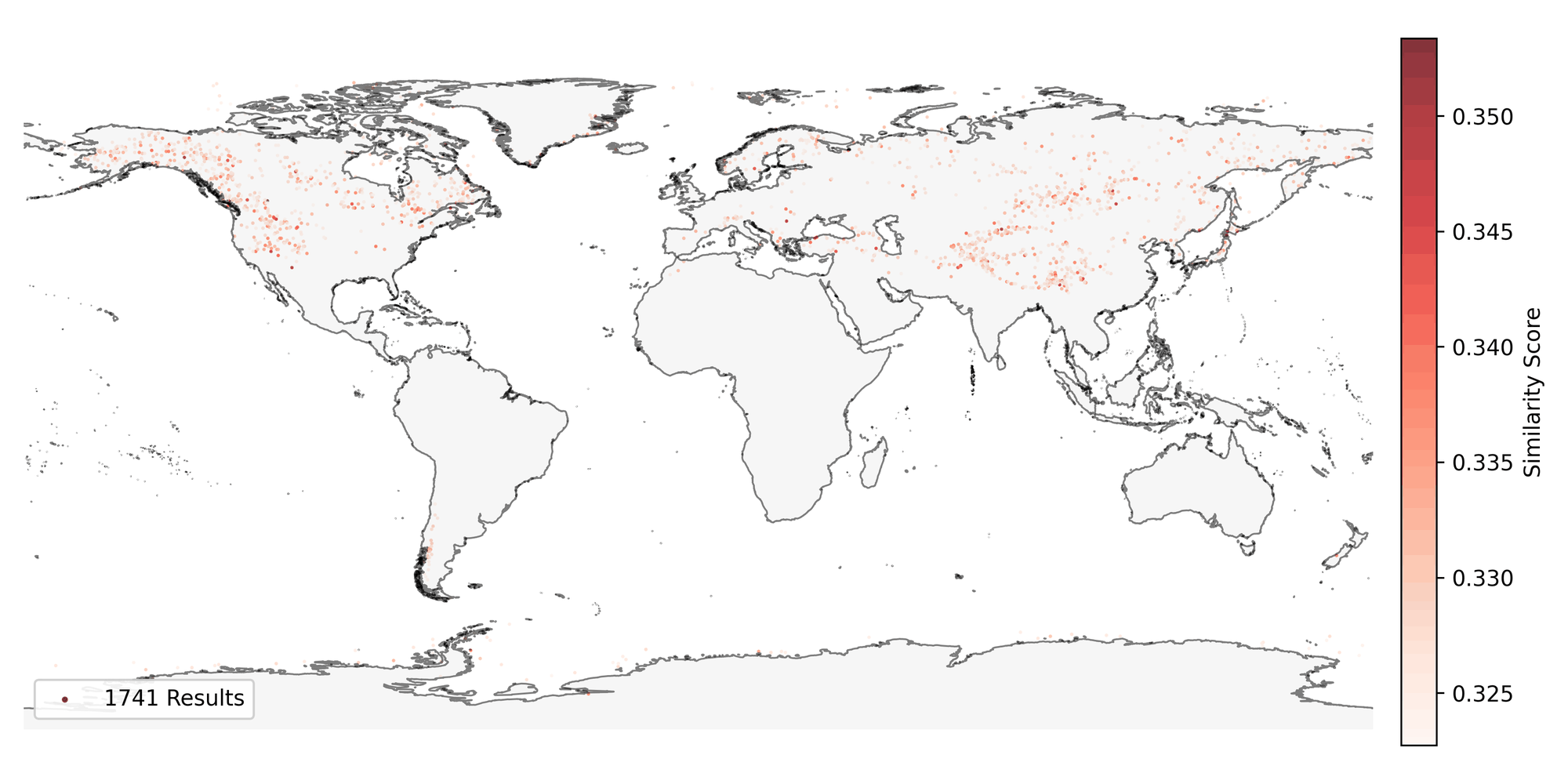}
        \caption{FarSLIP distribution}
    \end{subfigure}
    \hfill
    \begin{subfigure}{0.48\textwidth}
        \centering
        \includegraphics[width=\textwidth]{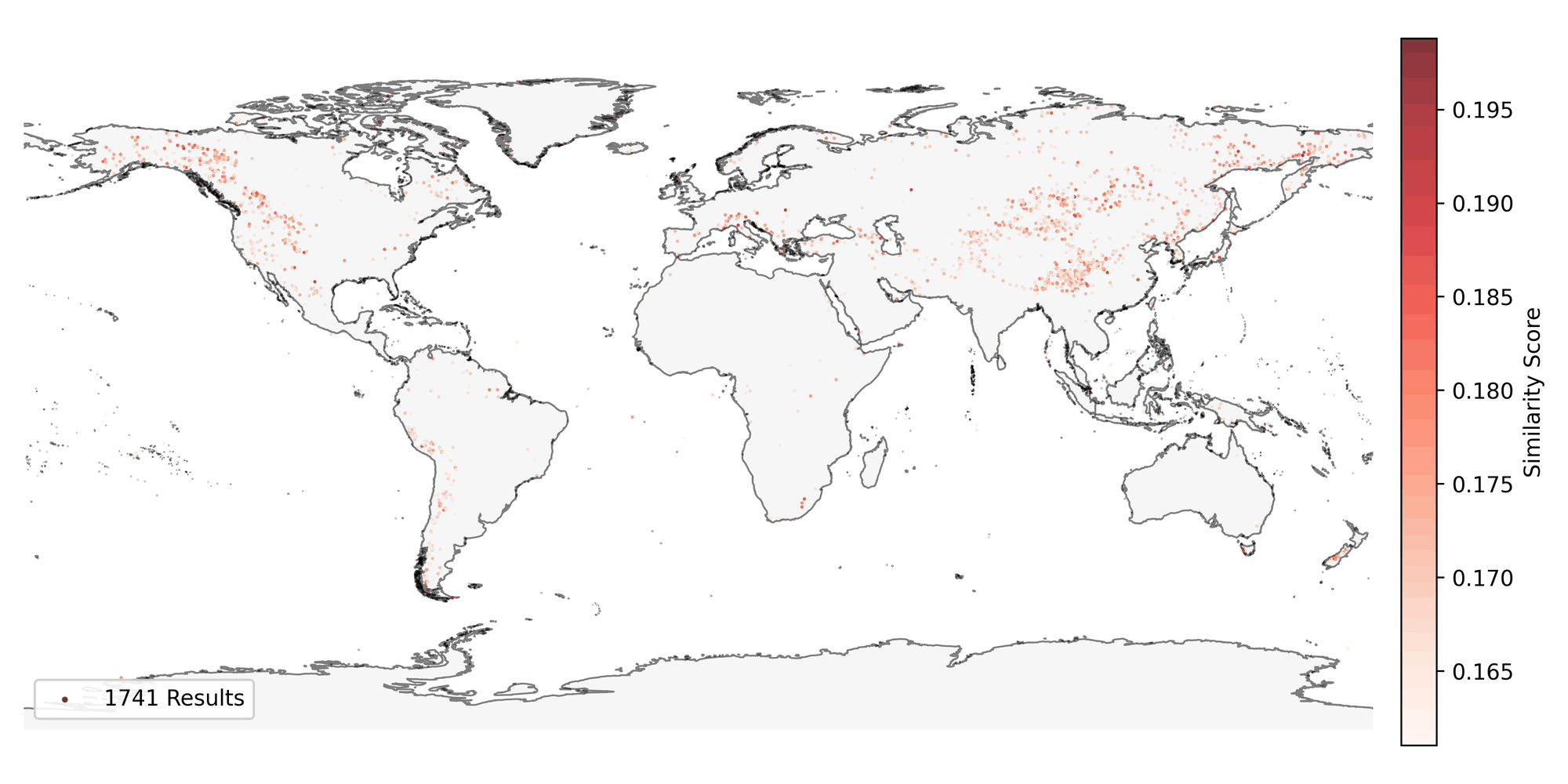}
        \caption{SigLIP distribution}
    \end{subfigure}
    \\
    \begin{subfigure}{0.48\textwidth}
        \centering
        \includegraphics[width=\textwidth]{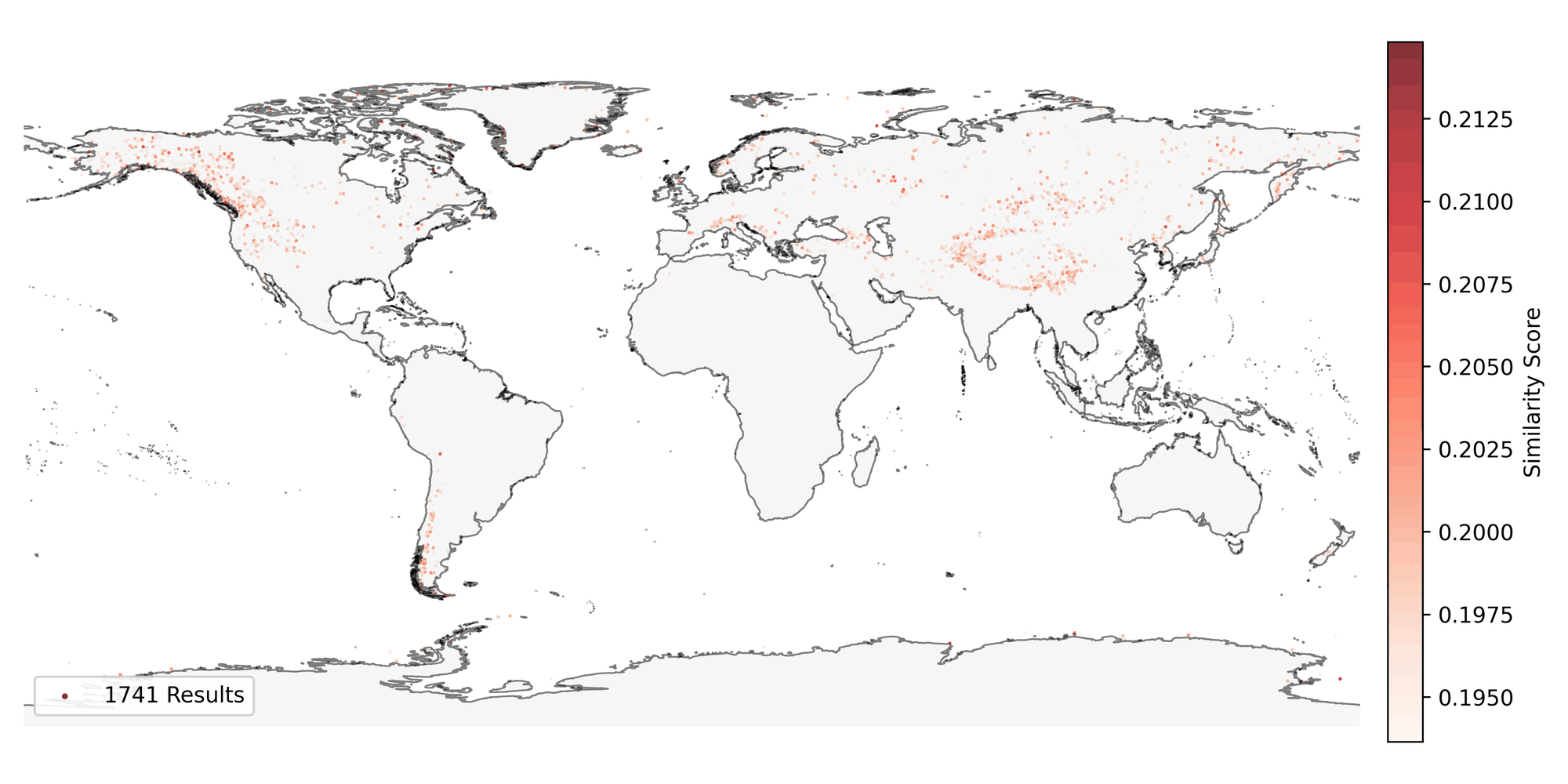}
        \caption{TIPSv2 distribution}
    \end{subfigure}
    \hfill
    \begin{subfigure}{0.48\textwidth}
        \centering
        \includegraphics[width=\textwidth]{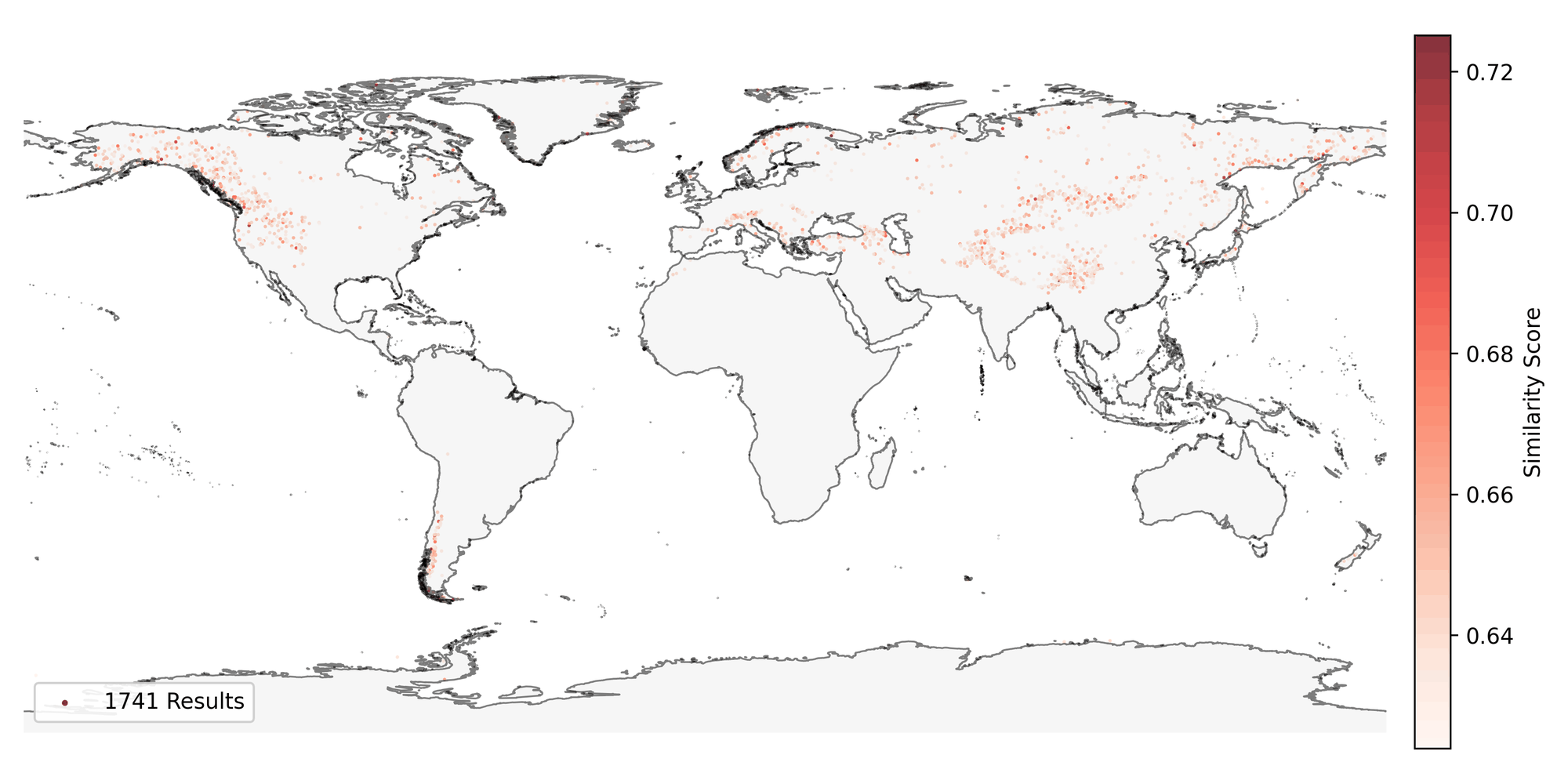}
        \caption{Qwen3-VL-Embedding distribution}
    \end{subfigure}
    \\
    \begin{subfigure}{0.48\textwidth}
        \centering
        \includegraphics[width=\textwidth]{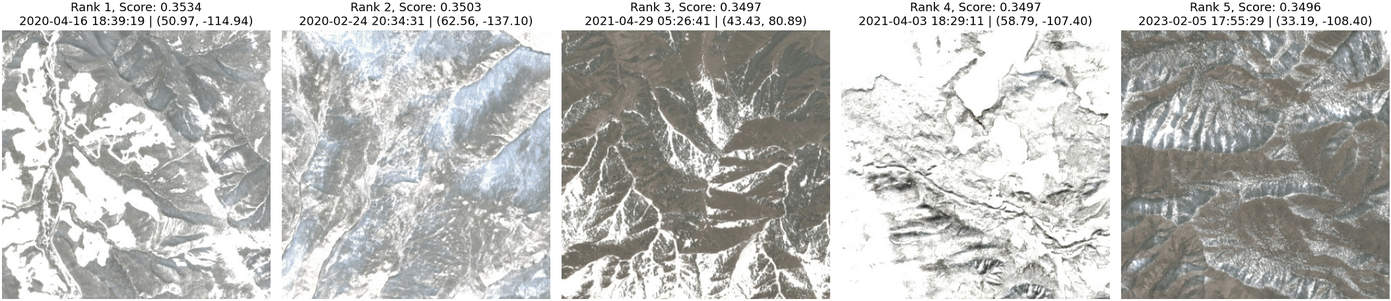}
        \caption{FarSLIP top-5}
    \end{subfigure}
    \hfill
    \begin{subfigure}{0.48\textwidth}
        \centering
        \includegraphics[width=\textwidth]{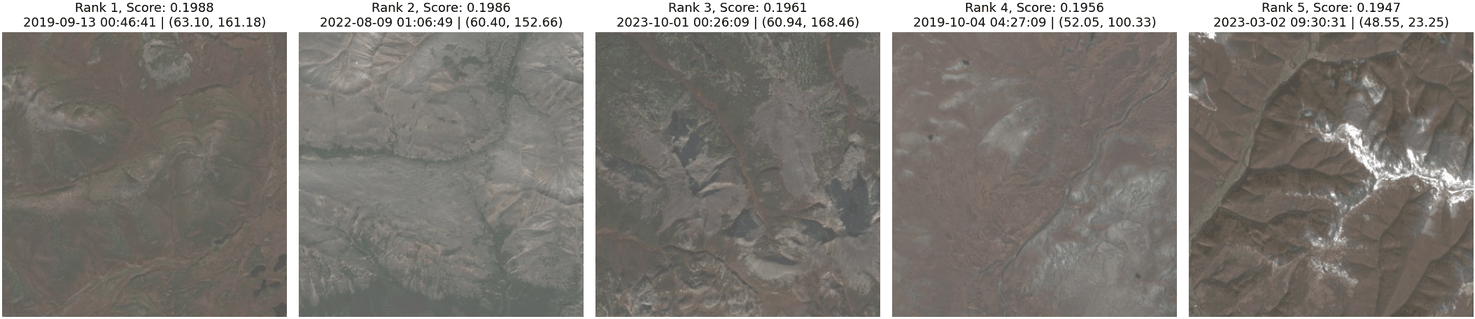}
        \caption{SigLIP top-5}
    \end{subfigure}
    \\
    \begin{subfigure}{0.48\textwidth}
        \centering
        \includegraphics[width=\textwidth]{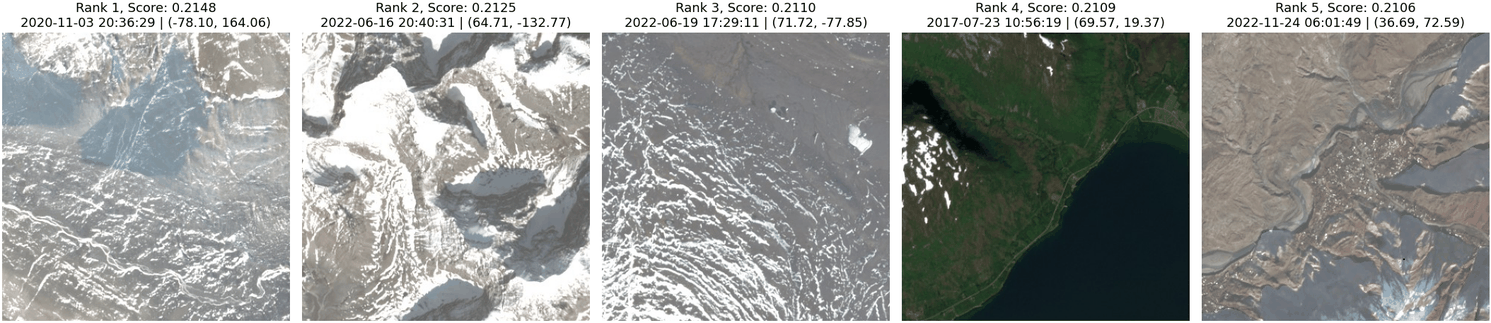}
        \caption{TIPSv2 top-5}
    \end{subfigure}
    \hfill
    \begin{subfigure}{0.48\textwidth}
        \centering
        \includegraphics[width=\textwidth]{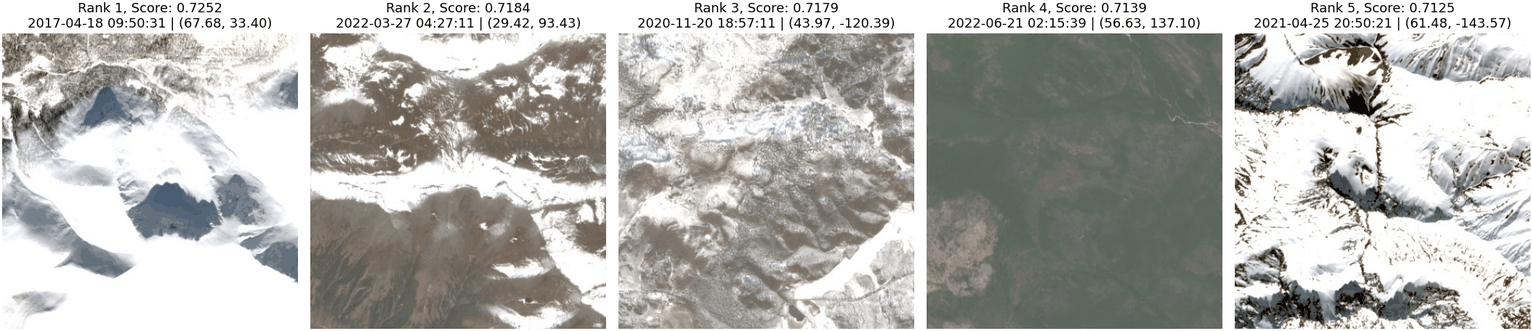}
        \caption{Qwen3-VL-Embedding top-5}
    \end{subfigure}
    \caption{Cross-model text-to-image retrieval for \texttt{a satellite image of snow covered mountains}. The first two rows show global distributions, and the last two rows show top-5 retrieved tiles.}
    \label{fig:snow}
\end{figure}

\clearpage
\paragraph{Natural feature: snow-covered mountains}
Figure~\ref{fig:snow} presents results for \texttt{a satellite image of snow covered mountains}. FarSLIP is strongly concentrated in the Northern Hemisphere, particularly across western North America and the major mountain belts of central and eastern Asia, while producing relatively few matches along the Andes and New Zealand's Southern Alps. In contrast, the more general web-scale or multimodal vision--language models---SigLIP, TIPSv2, and Qwen3-VL-Embedding---show a more balanced distribution across the two hemispheres, with clearer responses in both the Andes and Southern Alps as well as northern mountain systems. This contrast suggests a stronger northern geographic prior in FarSLIP for this prompt. The top-5 results provide a complementary view: FarSLIP consistently retrieves snow-covered mountains, Qwen3-VL-Embedding and TIPSv2 return several convincing examples with occasional weaker matches, whereas most SigLIP top-ranked tiles contain little visible snow.

\begin{figure}[H]
    \centering
    \begin{subfigure}{0.48\textwidth}
        \centering
        \includegraphics[width=\textwidth]{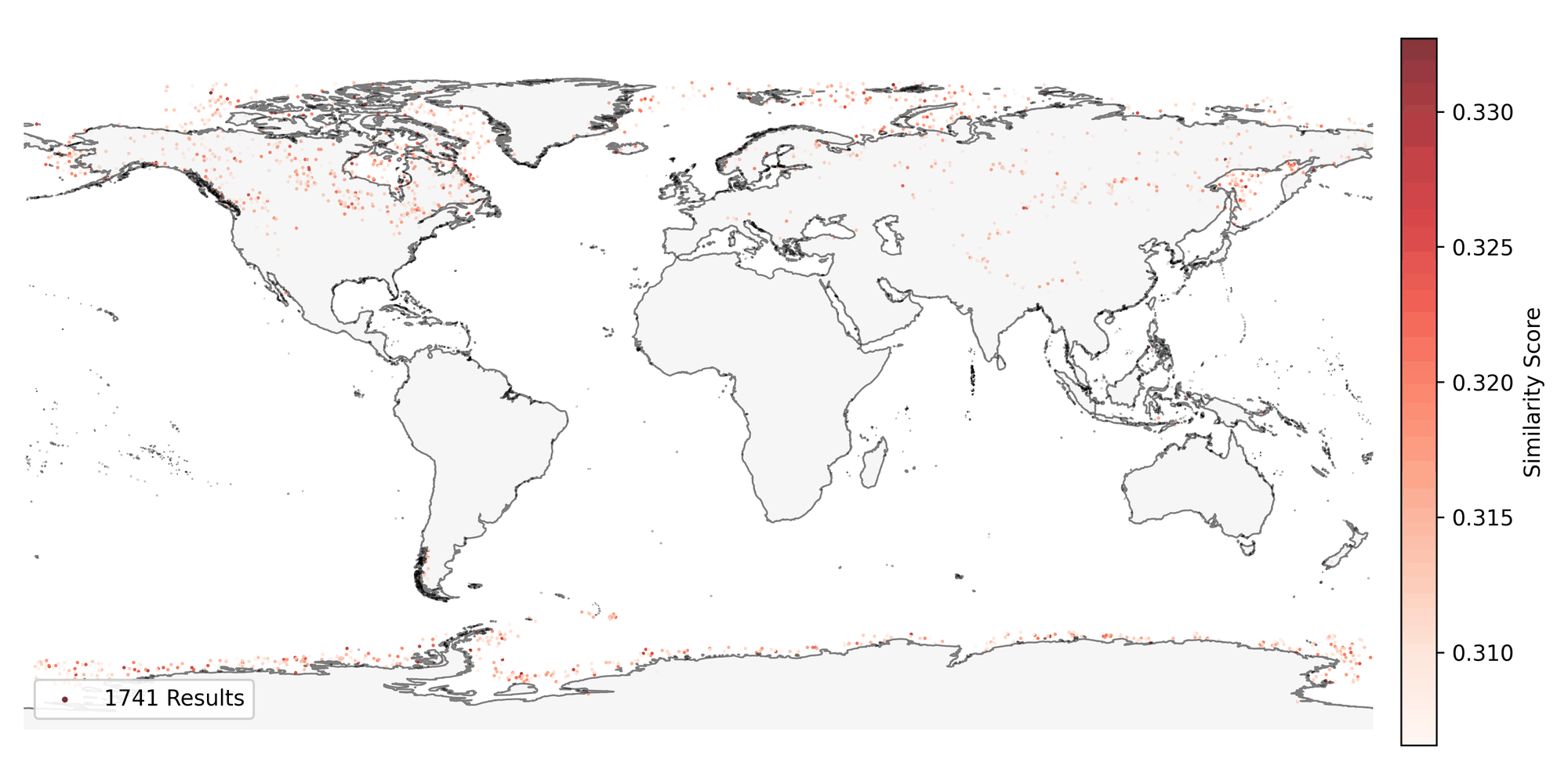}
        \caption{FarSLIP distribution}
    \end{subfigure}
    \hfill
    \begin{subfigure}{0.48\textwidth}
        \centering
        \includegraphics[width=\textwidth]{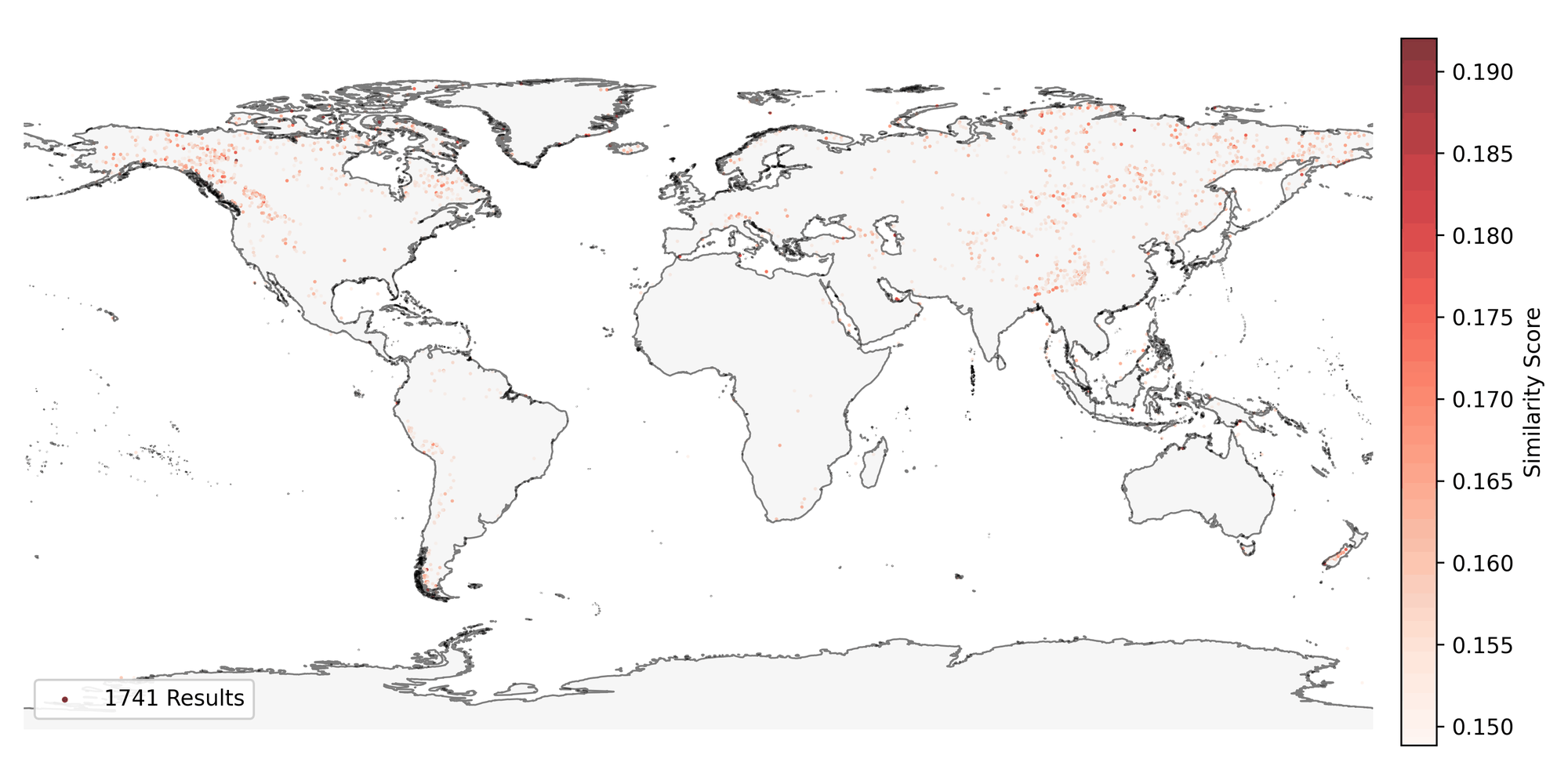}
        \caption{SigLIP distribution}
    \end{subfigure}
    \\
    \begin{subfigure}{0.48\textwidth}
        \centering
        \includegraphics[width=\textwidth]{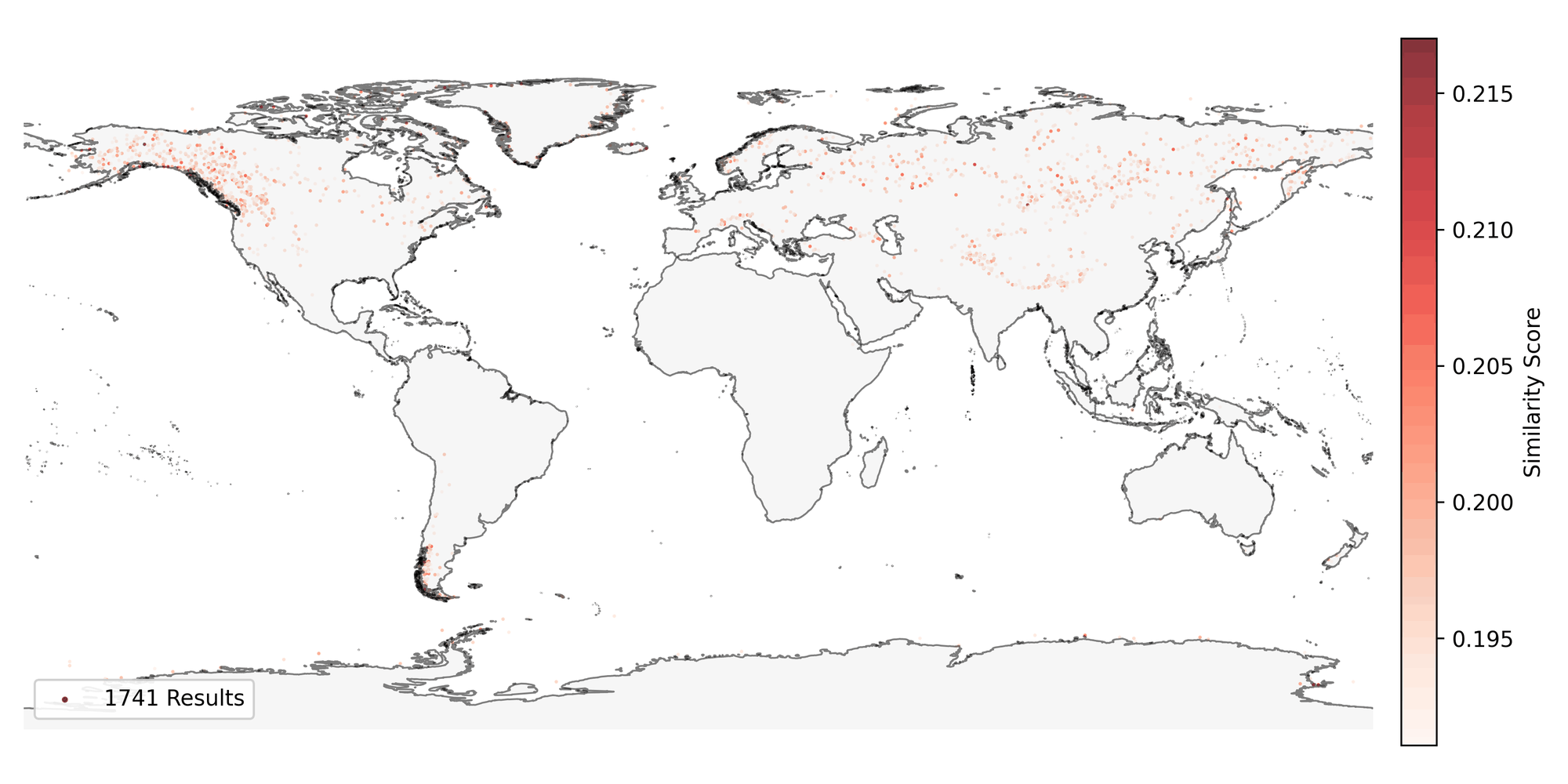}
        \caption{TIPSv2 distribution}
    \end{subfigure}
    \hfill
    \begin{subfigure}{0.48\textwidth}
        \centering
        \includegraphics[width=\textwidth]{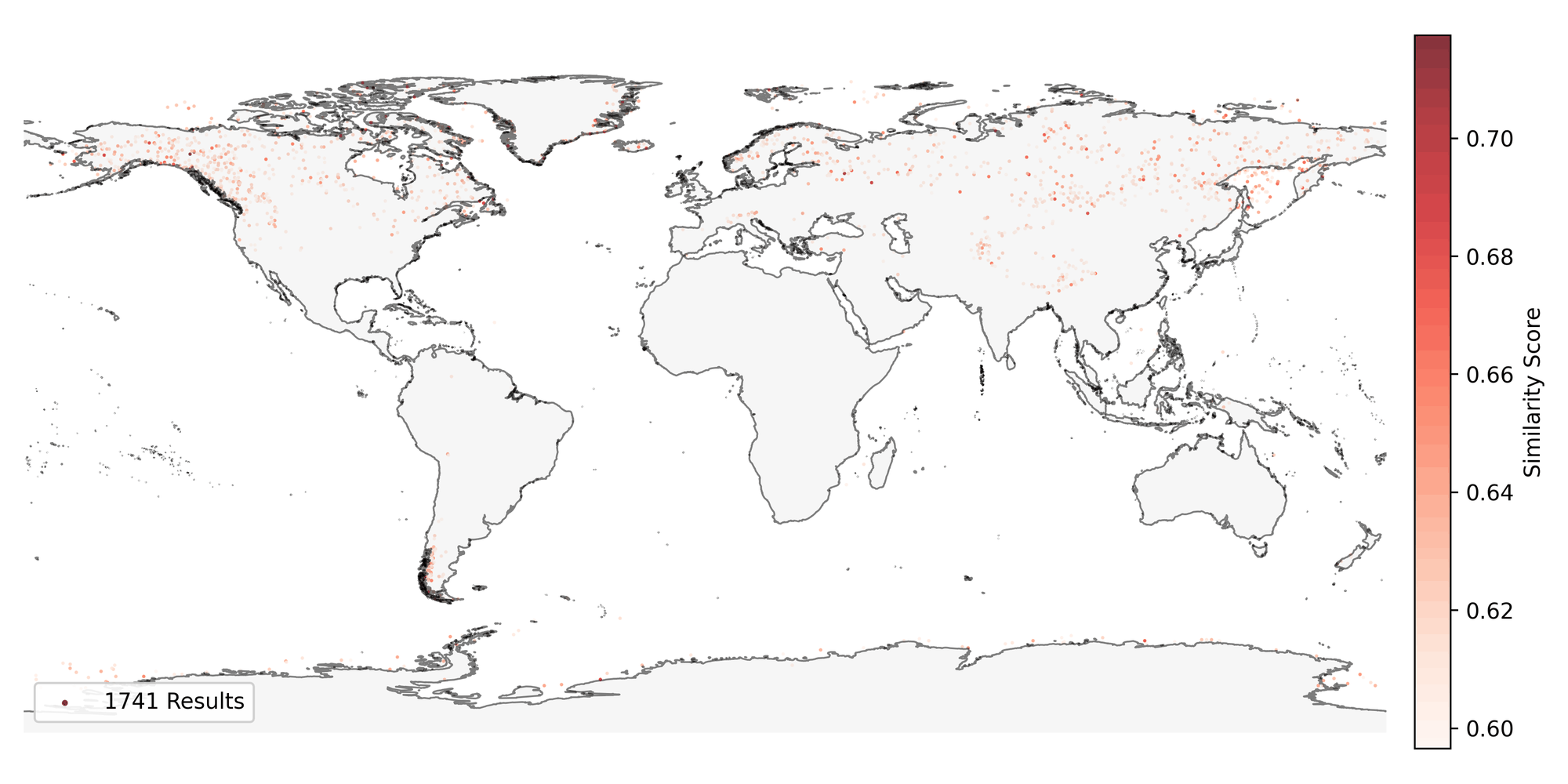}
        \caption{Qwen3-VL-Embedding distribution}
    \end{subfigure}
    \\
    \begin{subfigure}{0.48\textwidth}
        \centering
        \includegraphics[width=\textwidth]{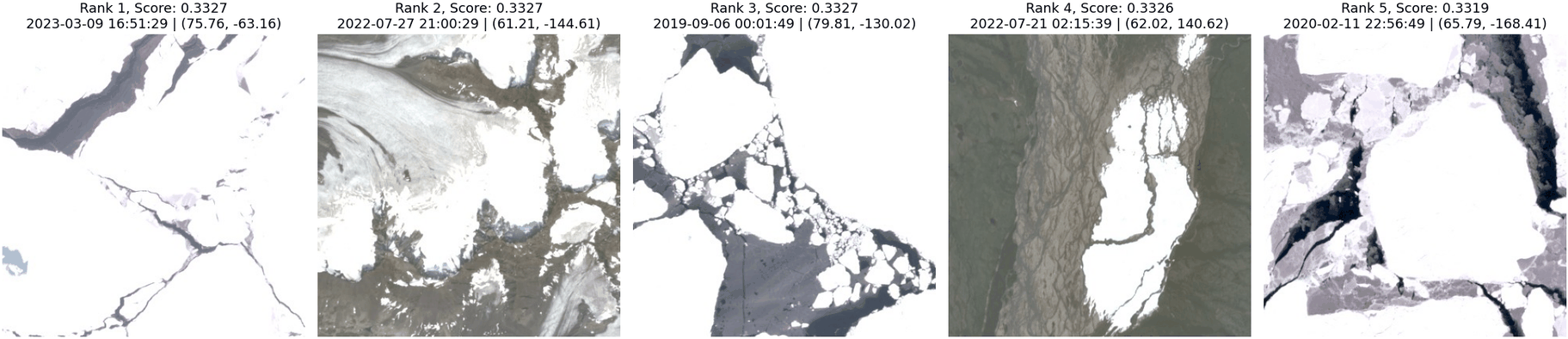}
        \caption{FarSLIP top-5}
    \end{subfigure}
    \hfill
    \begin{subfigure}{0.48\textwidth}
        \centering
        \includegraphics[width=\textwidth]{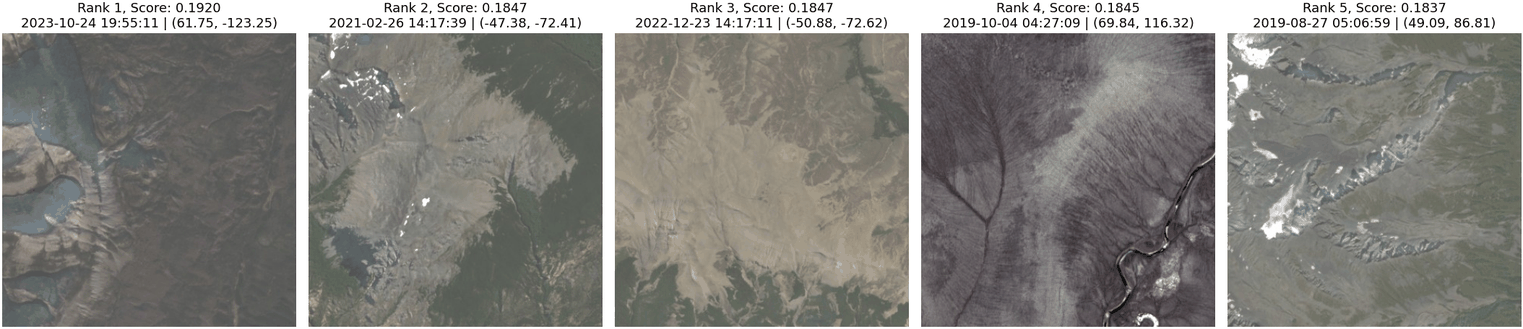}
        \caption{SigLIP top-5}
    \end{subfigure}
    \\
    \begin{subfigure}{0.48\textwidth}
        \centering
        \includegraphics[width=\textwidth]{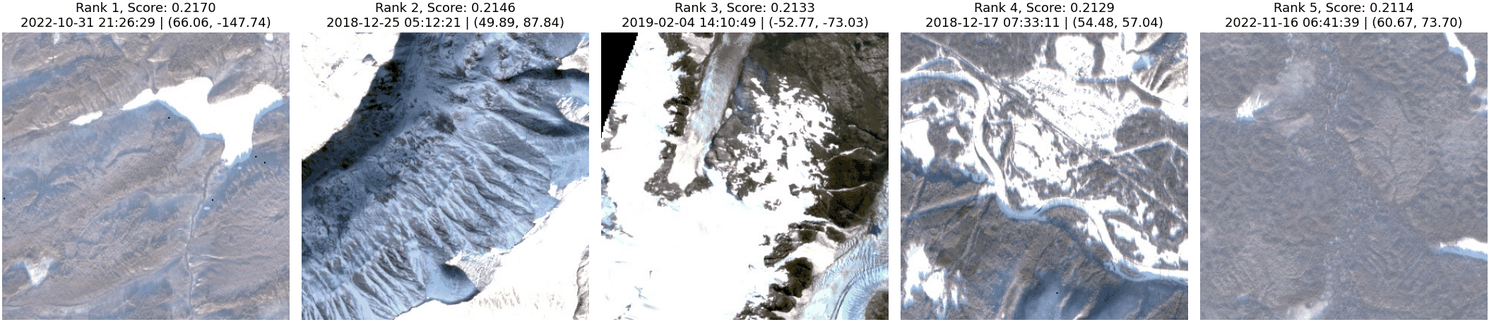}
        \caption{TIPSv2 top-5}
    \end{subfigure}
    \hfill
    \begin{subfigure}{0.48\textwidth}
        \centering
        \includegraphics[width=\textwidth]{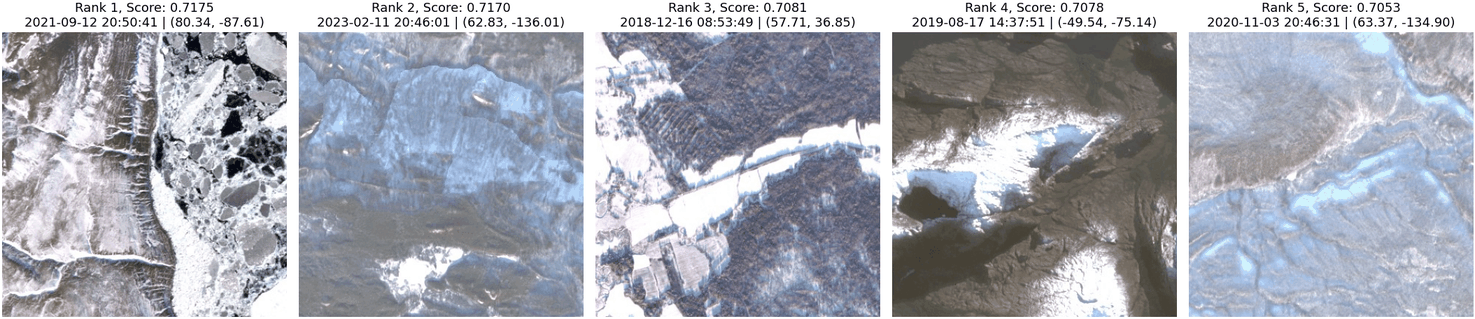}
        \caption{Qwen3-VL-Embedding top-5}
    \end{subfigure}
    \caption{Cross-model text-to-image retrieval for \texttt{a satellite image of a glacier}. The first two rows show global distributions, and the last two rows show top-5 retrieved tiles.}
    \label{fig:glacier}
\end{figure}

\clearpage
\paragraph{Natural feature: glacier}
Figure~\ref{fig:glacier} shows results for \texttt{a satellite image of a glacier}. FarSLIP is distinctive in assigning substantial similarity across the Antarctic margin, together with responses in Arctic and high-mountain regions. The other models concentrate more strongly on mountain belts associated with alpine glaciers and produce comparatively few matches over the continental ice sheets of Antarctica and Greenland. This difference suggests that FarSLIP associates the prompt with a broader polar-ice geography, whereas the general-purpose vision--language models emphasize mountainous glacier morphology. The top-5 results are consistent with this distinction: FarSLIP returns bright land ice, ice shelves, and sea ice; TIPSv2 and Qwen3-VL-Embedding retrieve several glacier or ice scenes but sometimes mix them with snow-covered mountains or frozen water; and most SigLIP results show barren mountain terrain with limited visible ice.

\clearpage
\subsection{Additional cross-model comparison for image-to-image retrieval}
We next compare image-to-image retrieval across heterogeneous encoders. Because this comparison includes many models, we report only the query and top-5 tiles rather than global distributions. Similarity scores are model-specific and are not compared across models.

\begin{figure}[H]
    \centering
    \begin{subfigure}{0.48\textwidth}
        \centering
        \includegraphics[width=\textwidth]{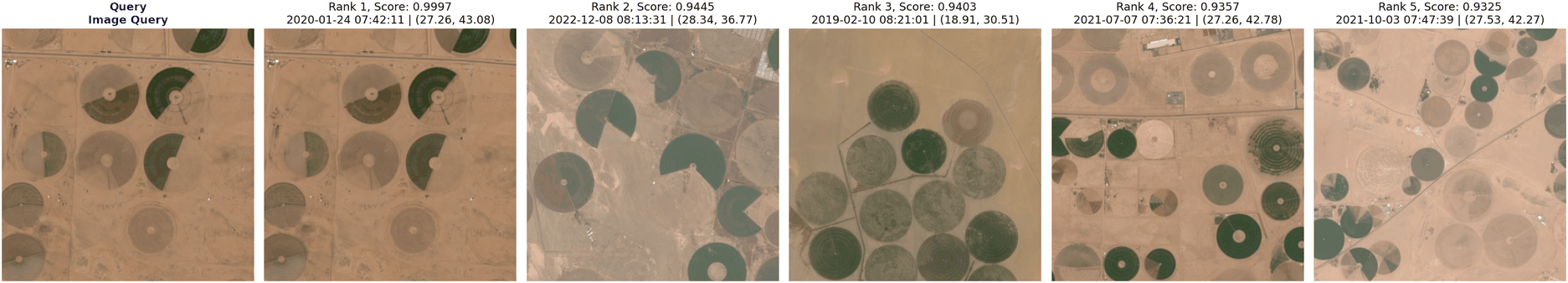}
        \caption{FarSLIP}
    \end{subfigure}
    \hfill
    \begin{subfigure}{0.48\textwidth}
        \centering
        \includegraphics[width=\textwidth]{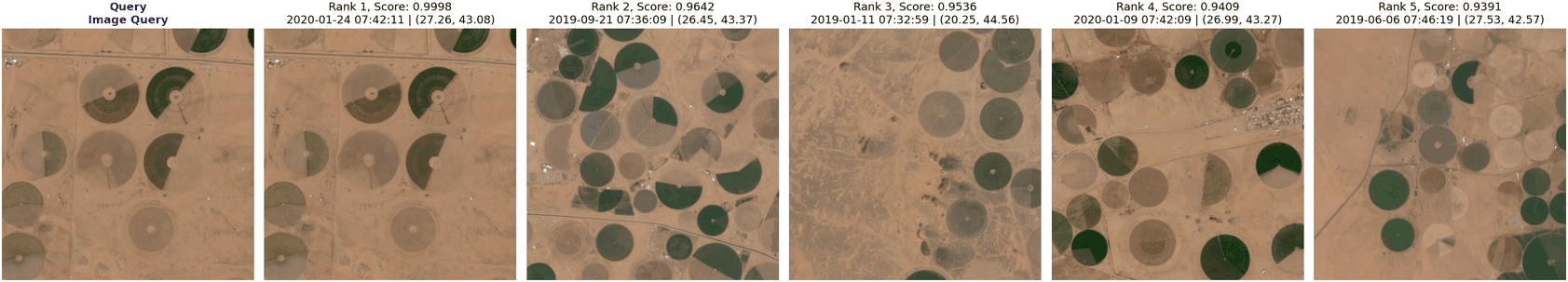}
        \caption{SigLIP}
    \end{subfigure}
    \\
    \begin{subfigure}{0.48\textwidth}
        \centering
        \includegraphics[width=\textwidth]{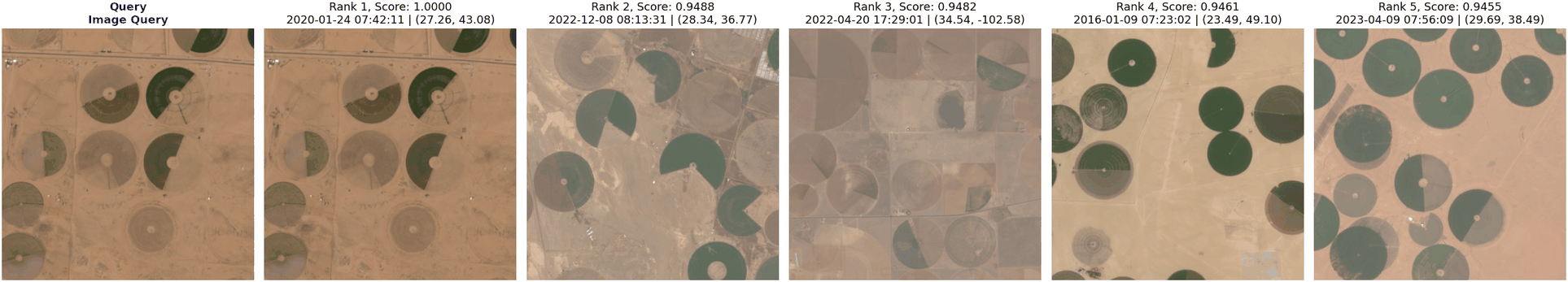}
        \caption{TIPSv2}
    \end{subfigure}
    \hfill
    \begin{subfigure}{0.48\textwidth}
        \centering
        \includegraphics[width=\textwidth]{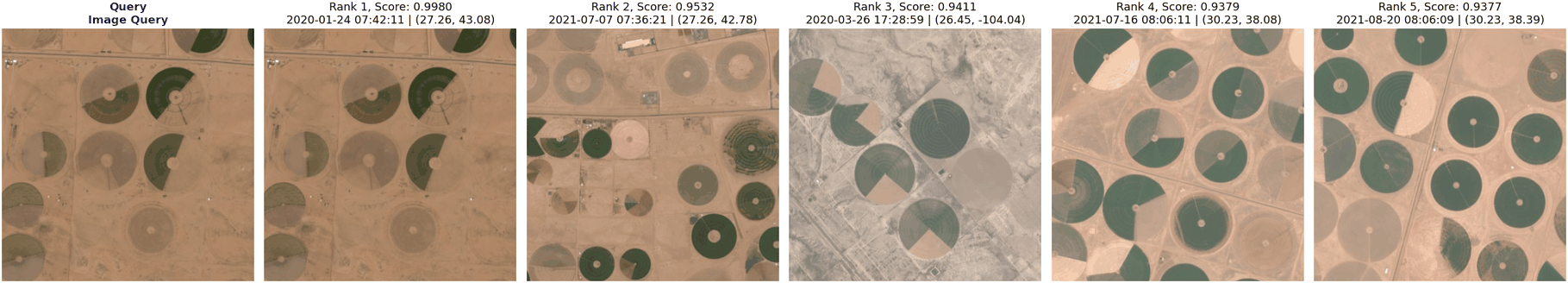}
        \caption{Qwen3-VL-Embedding}
    \end{subfigure}
    \\
    \begin{subfigure}{0.48\textwidth}
        \centering
        \includegraphics[width=\textwidth]{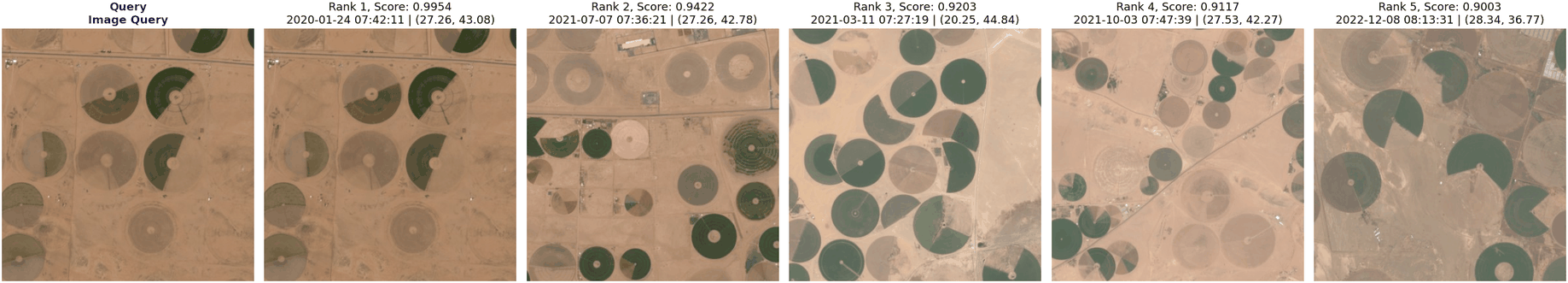}
        \caption{DINOv2}
    \end{subfigure}
    \hfill
    \begin{subfigure}{0.48\textwidth}
        \centering
        \includegraphics[width=\textwidth]{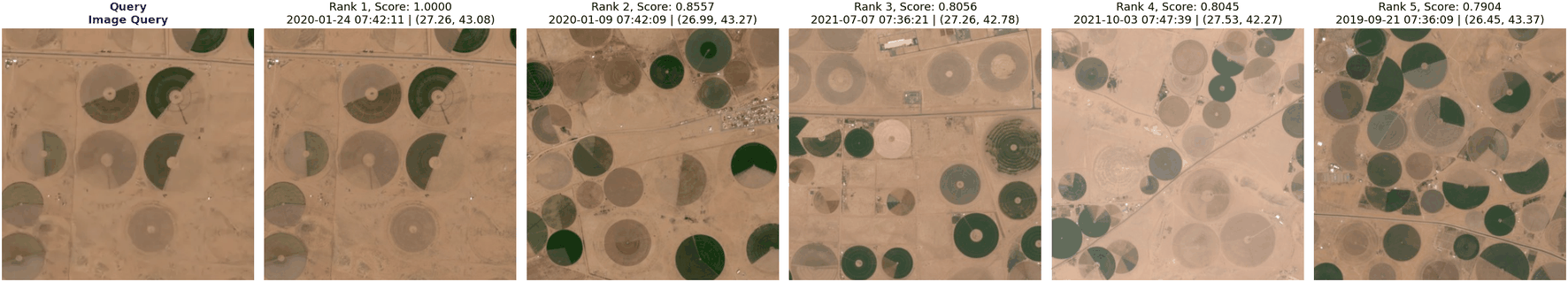}
        \caption{SatCLIP}
    \end{subfigure}
    \\
    \begin{subfigure}{0.48\textwidth}
        \centering
        \includegraphics[width=\textwidth]{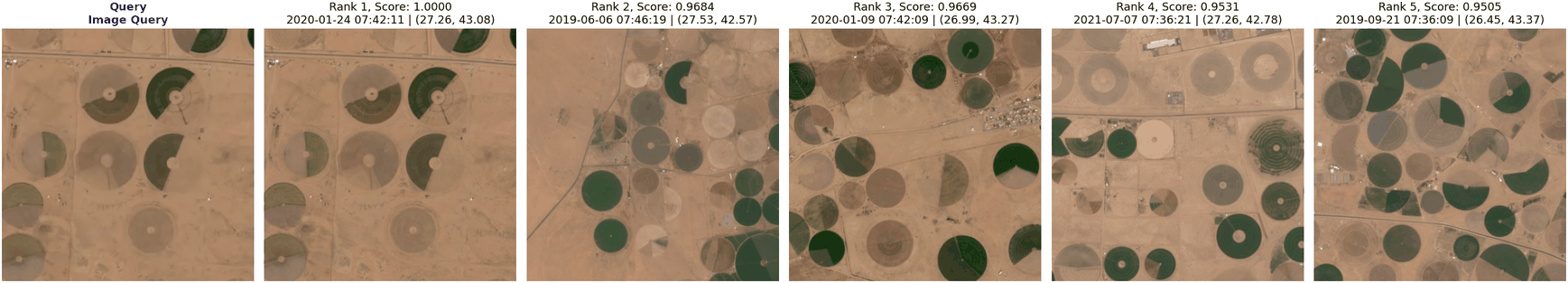}
        \caption{Clay}
    \end{subfigure}
    \hfill
    \begin{subfigure}{0.48\textwidth}
        \centering
        \includegraphics[width=\textwidth]{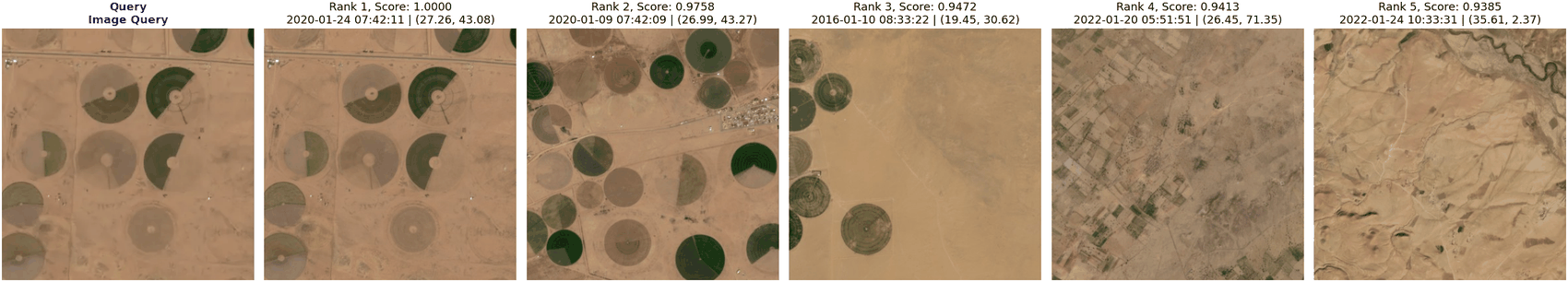}
        \caption{OlmoEarth v1.2}
    \end{subfigure}
    \caption{Image-to-image retrieval for a center-pivot farmland query across five RGB-based and three multispectral models. Each panel repeats the query at left and shows the model's top-5 results at right.}
    \label{fig:i2i_farmland}
\end{figure}

\paragraph{Center-pivot farmland query}
Figure~\ref{fig:i2i_farmland} uses a satellite tile containing center-pivot fields in an arid landscape as the query. All eight models return the query tile itself at rank~1, providing a consistent self-retrieval sanity check. The five RGB-based models preserve the defining circular irrigation geometry throughout their remaining top-ranked tiles: FarSLIP and TIPSv2 favor closely matching field shapes and crop states, while SigLIP, Qwen3-VL-Embedding, and DINOv2 retrieve a broader range of field densities and arrangements. Among the multispectral models, SatCLIP and Clay also consistently return center-pivot fields. OlmoEarth retrieves circular fields in its leading matches but shifts toward rectangular farmland and barren terrain at lower ranks, suggesting greater sensitivity to spectral, temporal, or background characteristics. Overall, the differences lie mainly in crop condition, field density, and scene context rather than in the dominant land-use pattern.

\begin{figure}[H]
    \centering
    \begin{subfigure}{0.48\textwidth}
        \centering
        \includegraphics[width=\textwidth]{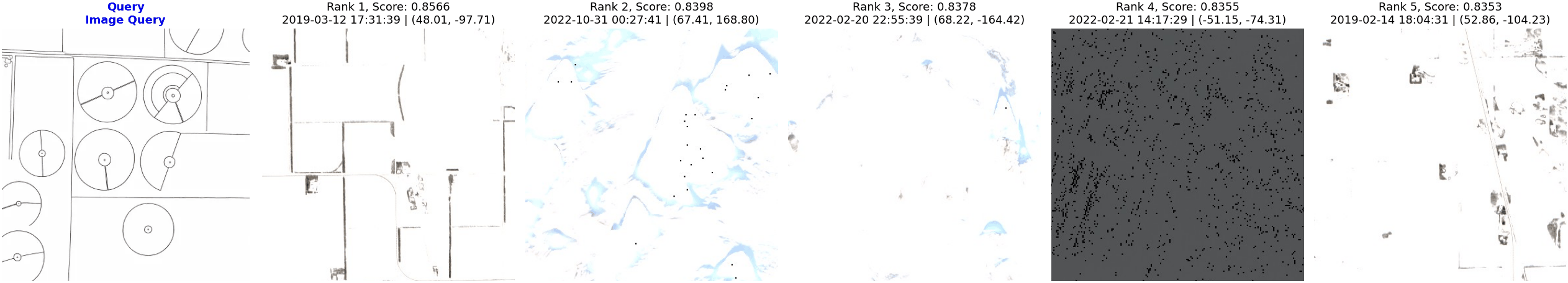}
        \caption{FarSLIP}
    \end{subfigure}
    \hfill
    \begin{subfigure}{0.48\textwidth}
        \centering
        \includegraphics[width=\textwidth]{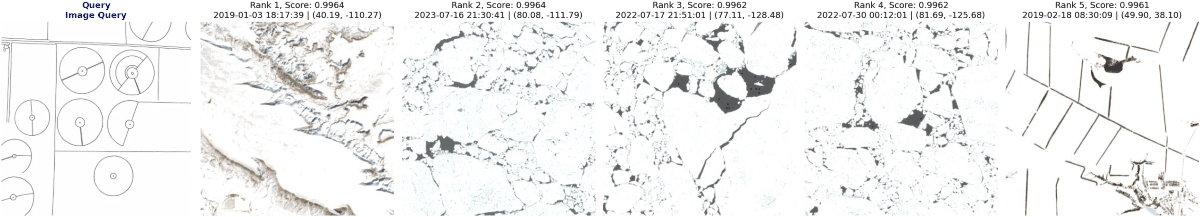}
        \caption{SigLIP}
    \end{subfigure}
    \\
    \begin{subfigure}{0.48\textwidth}
        \centering
        \includegraphics[width=\textwidth]{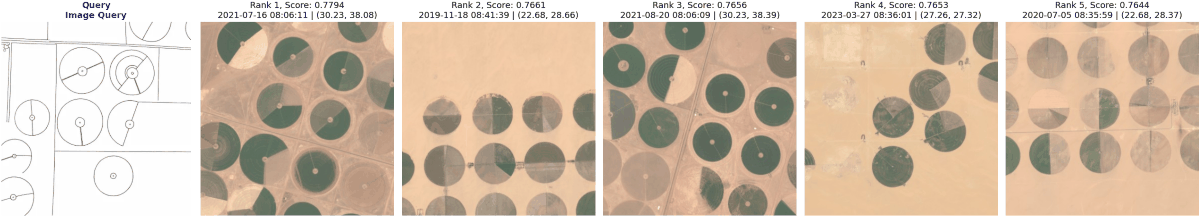}
        \caption{TIPSv2}
    \end{subfigure}
    \hfill
    \begin{subfigure}{0.48\textwidth}
        \centering
        \includegraphics[width=\textwidth]{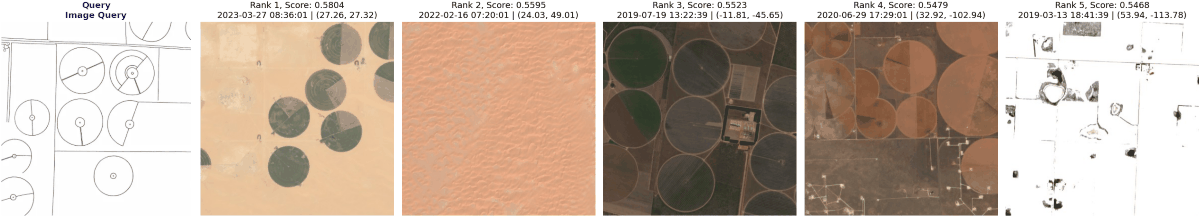}
        \caption{Qwen3-VL-Embedding}
    \end{subfigure}
    \\
    \begin{subfigure}{0.48\textwidth}
        \centering
        \includegraphics[width=\textwidth]{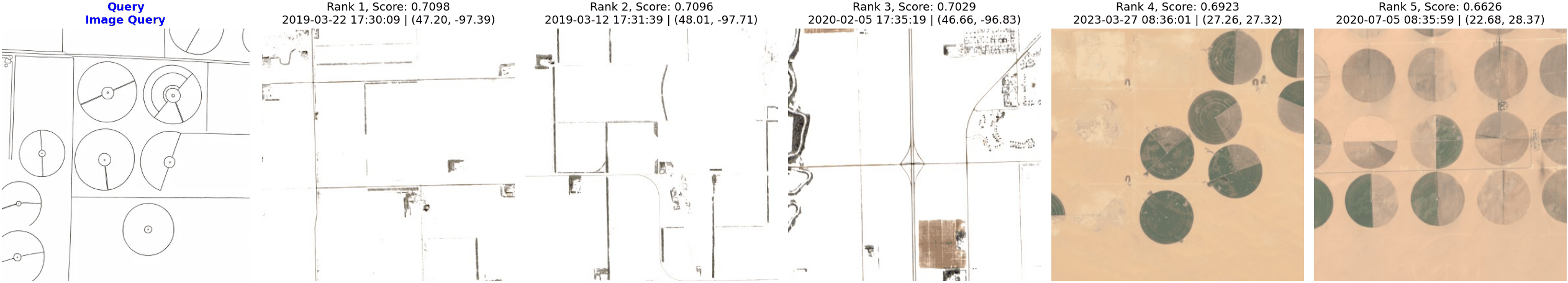}
        \caption{DINOv2}
    \end{subfigure}
    \caption{Image-to-image retrieval for a hand-drawn center-pivot-field sketch across five models. Each panel repeats the sketch at left and shows the model's top-5 results at right.}
    \label{fig:i2i_sketch}
\end{figure}

\clearpage
\paragraph{Hand-drawn sketch query}
To probe cross-domain generalization, we use a simple hand-drawn sketch of center-pivot fields and road boundaries as an intentionally out-of-domain query (Figure~\ref{fig:i2i_sketch}). TIPSv2 maps the sketch to the intended concept: all five results contain circular center-pivot fields. Qwen3-VL-Embedding retrieves several circular fields but also a sand-dune texture and a sparse high-contrast scene. DINOv2 and FarSLIP respond strongly to the sketch's thin dark lines and bright background, retrieving roads, parcel boundaries, or other linear patterns; DINOv2 additionally returns two center-pivot examples. SigLIP primarily retrieves snow or sea-ice scenes with dark cracks, suggesting that contrast and line texture dominate over the intended land-use semantics. The sketch therefore exposes whether a representation transfers across visual domains through semantic shape, low-level geometry, or surface texture.

% TODO: Upload the EarthEmbeddings to an anonymous dataset repo and create an anonymous space repo.

\end{document}